\documentclass{bmvc2k}

\title{Self-Supervised Learning of \\ Image Scale and Orientation}

\addauthor{Jongmin Lee}{ljm1121@postech.ac.kr}{1}
\addauthor{Yoonwoo Jeong}{jeongyw12382@postech.ac.kr}{1}
\addauthor{Minsu Cho}{mscho@postech.ac.kr}{1}

\addinstitution{
 Computer Vision Lab.\\
 POSTECH\\
 Pohang, Republic of Korea
}

\runninghead{Lee, Jeong, Cho}{Self-supervised Scale and Orientation}

\usepackage{amsmath}
\usepackage{amssymb}
\usepackage{gensymb}
\usepackage{multirow}
\usepackage{wrapfig}
\usepackage{tablefootnote}
\newcommand{\specialcell}[2][c]{%
  \begin{tabular}[#1]{@{}c@{}}#2\end{tabular}}

\usepackage{amsmath,amsfonts,bm}

\def\eqref#1{equation~\ref{#1}}

\def\1{\bm{1}}

\DeclareMathAlphabet{\mathsfit}{\encodingdefault}{\sfdefault}{m}{sl}
\SetMathAlphabet{\mathsfit}{bold}{\encodingdefault}{\sfdefault}{bx}{n}

\DeclareMathOperator*{\argmax}{arg\,max}

\def\eg{\emph{e.g}\bmvaOneDot}

\def\etal{\emph{et al}\bmvaOneDot}
\def\ie{\emph{i.e}\bmvaOneDot}

\begin{document}

\maketitle


\begin{abstract}

We study the problem of learning to assign a characteristic pose, i.e., scale and orientation, for an image region of interest. Despite its apparent simplicity, the problem is non-trivial; it is hard to obtain a large-scale set of image regions with explicit pose annotations that a model directly learns from. To tackle the issue, we propose a self-supervised learning framework with a histogram alignment technique. It generates pairs of image patches by random  rescaling/rotating and then train an estimator to predict their scale/orientation values so that their relative difference is consistent with the rescaling/rotating used. 
The estimator learns to predict a non-parametric histogram distribution of scale/orientation without any supervision. Experiments show that it  significantly outperforms previous methods in scale/orientation estimation and also improves image matching and 6 DoF camera pose estimation by incorporating our patch poses into a matching process.
Our code is available on \url{https://github.com/bluedream1121/self-sca-ori}.

\end{abstract}



\begin{wrapfigure}{R}{0.49\textwidth}
    \centering
    \scalebox{0.56}{
    \includegraphics[]{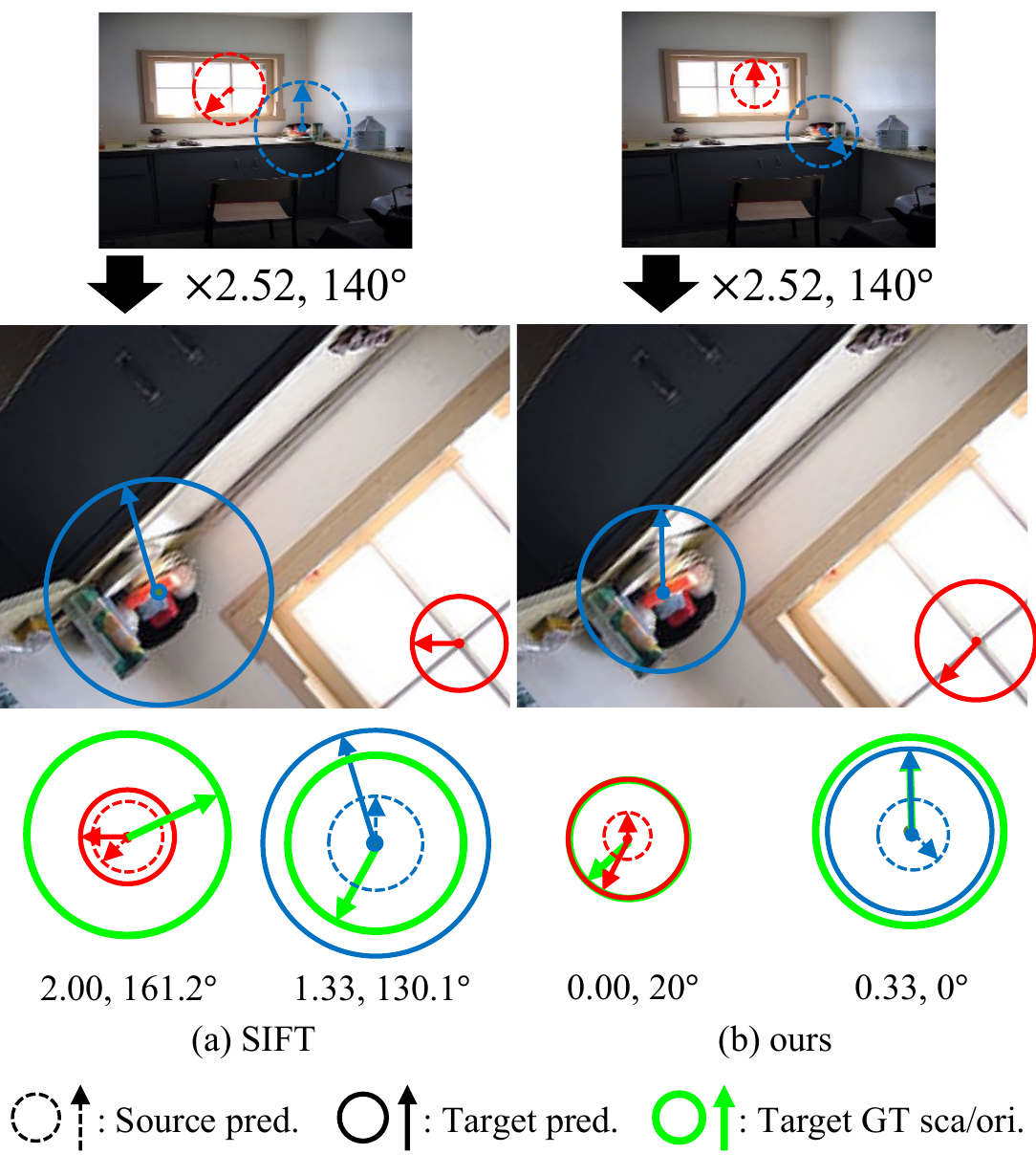}}
    \vspace{-0.2cm}
    \caption{
    Comparison of scale/orientation estimation of SIFT~\cite{lowe2004distinctive} vs. ours. 
    The size of circles represents the scale, and the direction of arrows means the orientation.
    At the bottom, the green circle/arrow depicts the true scale/orientation given the estimation of the top and  
    the numbers mean the errors in relative scale/orientation. 
    } 
    \label{fig:teaser}  
\end{wrapfigure}


\section{Introduction}

Local feature representation lies at the heart of computer vision, and extensive research has been conducted 
on detecting and/or describing local features~~\cite{bay2006surf, calonder2010brief, lowe2004distinctive, rublee2011orb}. With the remarkable advance of convolutional neural networks (CNNs) ~\cite{he2016deep, simonyan2014very, tian2017l2}, 
the dense feature map output of convolutional layers has largely replaced the classic hand-crafted feature representation in a wide range of tasks~\cite{detone2018superpoint, dusmanu2019d2, revaud2019r2d2, barroso2019key, rocco2020ncnet, lee2021learning}.
However, since the convolutional feature map is equivariant only to translation but not to the other common pose variations, \eg, scaling and rotating, 
 assigning a characteristic pose of an image or region of interest is required to extract an accurate descriptor for many vision problems such as visual correspondence, registration, retrieval, localization, and 3D reconstruction~\cite{balntas2017hpatches, long2014convnets, lowe2004distinctive,  noh2017large, radenovic2018revisiting, sattler2018benchmarking, schonberger2016structure, schoenberger2016mvs}; \eg, the characteristic scale and/or orientation can be used to extract pose-normalized features from images with different viewpoints or object poses. 
 
Despite its apparent simplicity, the problem of learning to assign a characteristic pose, \ie, scale and orientation, for an image region is non-trivial; it is hard to obtain a large-scale set of image regions with explicit pose annotations that a model directly learns from.
To tackle the issue, 
recent methods~\cite{ono2018lf,shen2019rf, yi2016lift, yi2016learning, mishkin2018repeatability} use an implicit learning approach with a surrogate objective where they treat scale and/or orientation as a latent variable; 
they indirectly train a pose regressor by maximizing the similarity between image regions that are aligned using the estimated pose values. 

In this paper, we propose a self-supervised explicit learning framework via a histogram alignment technique. 
Instead of implicit learning with a surrogate objective~\cite{yi2016learning, ono2018lf, mishkin2018repeatability, shen2019rf}, we generate self-supervised pairs of image regions by random scaling and rotating and then train a model to predict pose value distributions so that their relative difference is consistent with scaling and rotating being used. In contrast to the previous learning-based methods, we advocate the histogram output for pose, which is similar to SIFT~\cite{lowe2004distinctive}, and propose a histogram alignment technique for self-supervised learning. The method learns a non-parametric and multi-modal distributions of scale and orientation without any human annotations, effectively resolving the challenge of defining and annotating characteristic poses for image regions.   
Experimental results show a significant improvement over the previous method both in scale and orientation estimation on the proposed PatchPose dataset and the HPatches~\cite{balntas2017hpatches} dataset, demonstrating the effectiveness of our self-supervised learning framework. 
Moreover, the image matching result on HPatches~\cite{balntas2017hpatches} shows the patch extraction effect to mean matching accuracy (MMA) using our method. The 6 DoF pose estimation results on IMC2021~\cite{jin2021image} show the outlier rejection effect by our scale and orientation.
The code and models are publicly available at \href{https://github.com/bluedream1121/self-sca-ori}{[this link]}.
\vspace{-0.4cm}
\section{Related work}

\smallbreak 
\noindent \textbf{Scale and orientation estimation.}
The most representative is the scale-invariant feature transform (SIFT)~\cite{lowe2004distinctive}, where Lowe introduces gradient histograms for orientation estimation and difference of Gaussians for scale estimation.
Despite its success, it often fails when geometric or photometric deformation is present. 
Bay~\etal~\cite{bay2006surf} improve SIFT by Hessian-based descriptors and integral images.
Rublee~\etal~\cite{rublee2011orb} propose efficient measure of corner orientation using intensity centroid~\cite{rosin1999measuring} on the FAST detectors~\cite{rosten2006machine}. These classical methods use handcrafted algorithms to obtain scale and orientation without learning. 
Recent research has investigated learning-based methods to estimate characteristic scale and/or orientation for image patches. 
Yi~\etal~\cite{yi2016learning} introduce a CNN that learns to predict the characteristic orientation of an image patch. To avoid the difficulty of defining the characteristic orientation, they train the CNN by minimizing the distance between orientation-normalized descriptors of two matchable patches.
In the subsequent work of~\cite{ono2018lf, shen2019rf}, they integrate scale/orientation estimation with feature point detection and description for image matching.
More recent studies~\cite{barroso2019key, ebel2019beyond, liu2019gift, pautrat2020online} aim to extract local descriptors that are invariant or covariant with respect to geometric variations within a local region.
The aforementioned learning-based methods all share common strategies: (1) regression-based estimation, (2) implicit learning by improving descriptor matching, and (3) the use of matchable pairs obtained from different datasets, e.g. phototourism~\cite{  mikolajczyk2005performance, thomee2016yfcc100m, wilson2014robust}, ScanNet~\cite{dai2017scannet} with depth information and HPatches~\cite{balntas2017hpatches} with ground-truth homography. 
In contrast, our method uses (1) histogram-based estimation, (2) self-supervised explicit learning, and (3) unsupervised datasets with random transformation.

There also exists previous work on estimating more general transformation beyond scale and orientation. For example, Mikolajczyk and Schmid~\cite{mikolajczyk2004scale} introduce a scale/affine-invariant keypoints detector using an affine shape estimator based on the second moment matrix.
Mishkin~\etal~\cite{mishkin2018repeatability} propose to learn an affine-covariant region detector using the spatial transformer network~\cite{jaderberg2015spatial} and the triplet margin loss. In most cases of current image matching applications, however, the use of scale and orientation only is still dominant. 


\smallbreak  
\noindent \textbf{Self-supervised learning.}
The task of orientation prediction has often been used as a pretext task for self-supervised representation learning.
Gidaris~\etal~\cite{gidaris2018unsupervised} introduce a classification task of predicting a rotated angle of an image for representation learning.
Feng~\etal{}~\cite{feng2019self} propose to decouple the rotation discrimination from instance discrimination. 
While learning to estimate image orientation in a self-supervised manner, 
the predicted orientation from these methods cannot be used for the characteristic orientation we consider in this work; they assume a fixed and predefined canonical orientation (\ie, upright) for each object class and simply predict the rotation from it by learning the class information, which cannot generalize to arbitrary images to be aligned.
In contrast, our method does not assume any prior information about predefined object classes and their canonical orientations. 
\smallbreak  
\noindent \textbf{Invariant feature learning.}
Our approach to estimating characteristic scale and orientation is also relevant to learning image relations for invariant feature representation~\cite{memisevic2010learning, memisevic2012multi, sohn2012learning}.
Memisevic and Hinton~\cite{memisevic2010learning} approximate a three-dimensional interaction tensor of a higher-order Boltzmann machine via factorizing the tensor. They investigate how image transformations affect the filters of the proposed model in a visual analogy task. Memisevic~\cite{memisevic2012multi} proposes a conservative detector, called the subspace rotation detector, which generates a content-independent representation. 
Sohn and Lee~\cite{sohn2012learning} extend the RBM to capture transformation between eigenfeatures of two images. With the predicted transformation, their model extracts transformation-invariant features, which are beneficial in image classification.


The contribution of this paper is three-fold. First, we introduce a self-supervised learning framework to estimate the characteristic scale and orientation for an arbitrary image patch.
Second, we propose a histogram alignment technique for learning to estimate multi-modal distribution of scale/orientation. 
Third, experimental evaluation on scale/orientation estimation benchmarks demonstrates the effectiveness of our approach, significantly outperforming recent methods.  
\vspace{-0.4cm}
\begin{figure}[t]
    \centering
    \scalebox{0.7}{
    \includegraphics{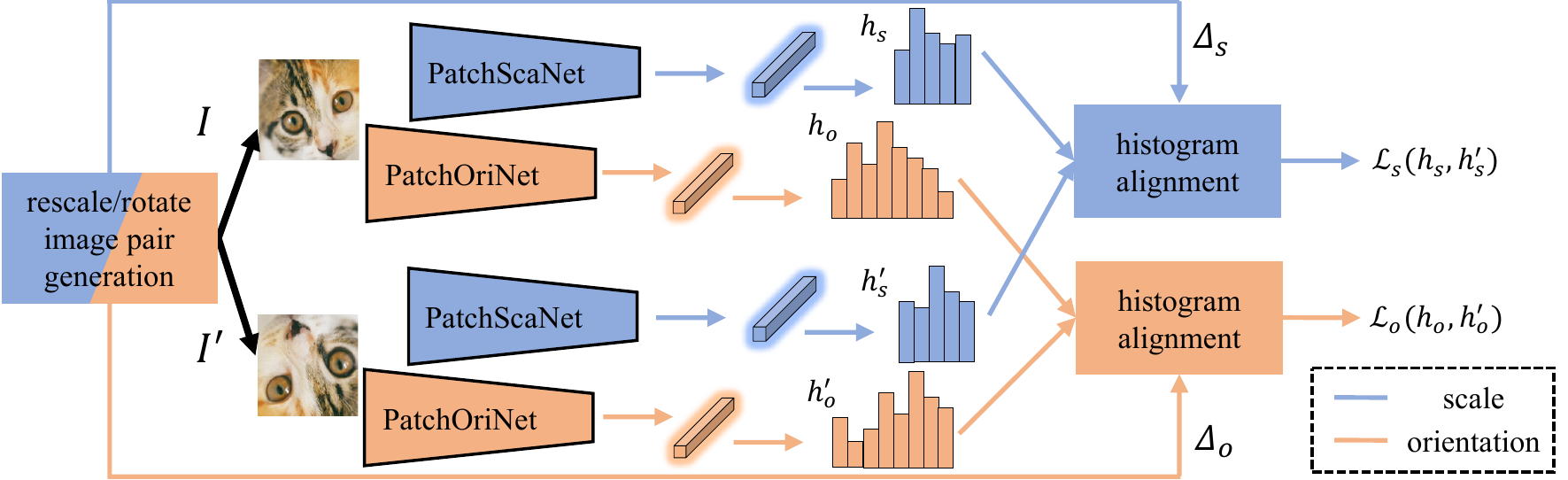}
    }
    \vspace{-0.2cm}
    \caption{Overview of our self-supervised framework for learning a patch pose, \ie, orientation and scale. 
    Given a pair of image patches with rescaling/rotating, we feed them to the patch pose estimation networks that output scale/orientation histograms for each image patch. We compare the two histograms by the histogram alignment technique and compute the loss, which is used for training the networks via backpropagation. 
    } 
    \label{fig:overall_arch}
\end{figure}
\section{Method}
In this section, we introduce the patch pose estimation network that learns to predict characteristic scale and orientation of an image patch. 
We first describe the model architecture and then explain our strategy for self-supervised learning. 
\vspace{-0.2cm}
\subsection{Patch pose estimation networks}~\label{sec:network_architecture}
The patch pose estimation networks are designed to predict the characteristic pose, \ie, orientation and scale, of a given image patch. 
We cast the patch pose estimation into the problem of predicting a probability distribution over candidate pose values rather than that of regressing a target pose value. 
The basic form of the architecture thus consists of a convolutional feature extractor followed by MLPs with softmax output that produces a histogram over a set of candidate pose values: 
\begin{equation}
    h = \sigma( \texttt{MLP}(\texttt{CONV}(I)) )
\end{equation}
where $\sigma(\cdot)$ is the softmax function and $h \in {\{x \in \mathbb{R}:0 \leq x \leq 1\}}^{B}$ is the histogram distribution of pose with $B$ bins, \ie, either orientation or scale. 
Figure~\ref{fig:overall_arch} shows the overall architecture of our model.

\begin{wrapfigure}{R}{0.53\textwidth}
    \centering
    \scalebox{0.45}{
    \includegraphics{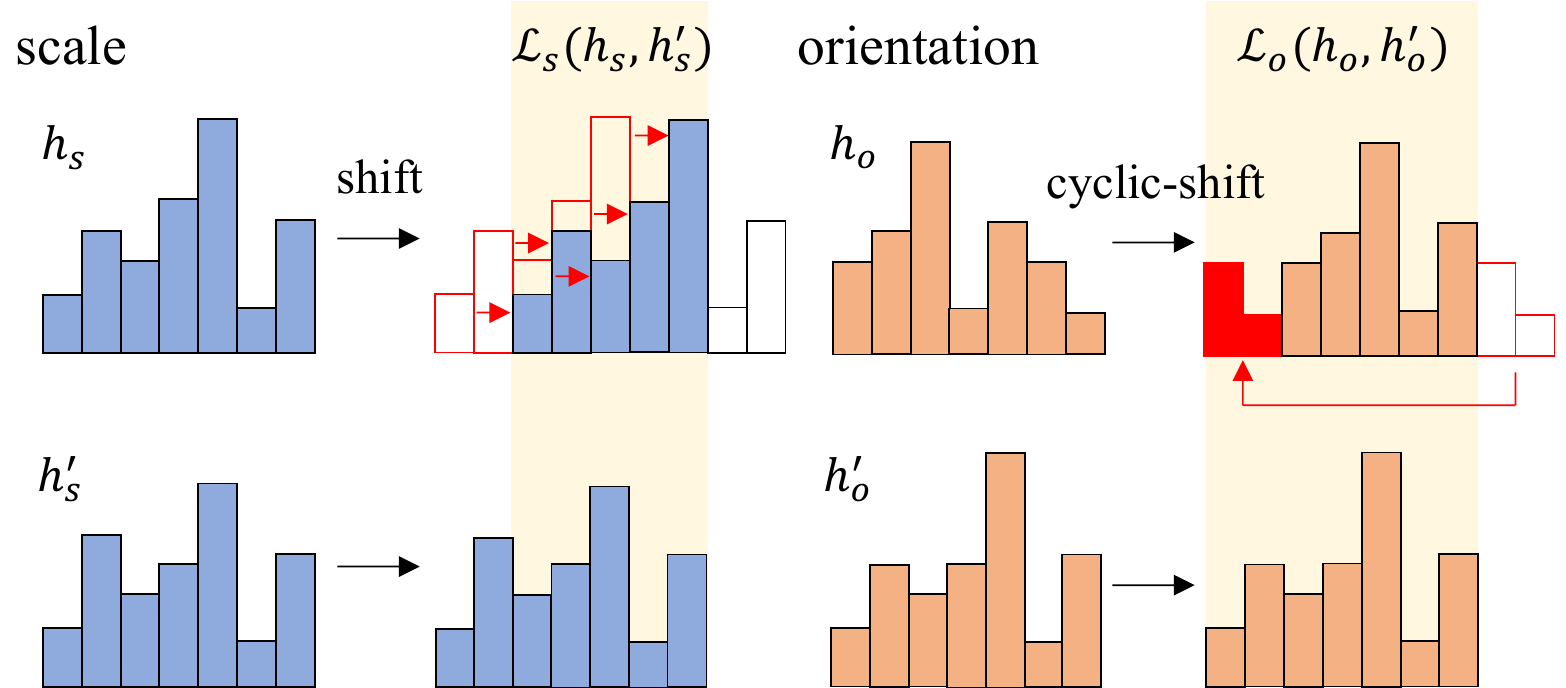}
    }
     \vspace{-0.6cm}
    \caption{Histogram alignment for scale and orientation. The histogram $h$ is shifted by a ground-truth $\Delta$. 
    For scale, the overlapping regions of the shifted one and the other are used to compute the loss. 
    For orientation, the shift operation is circular so that the entire regions of histograms are used.
    }
    \label{fig:histogram_alignment}
\end{wrapfigure}

The output histogram $h$, \ie, $B$ bins of discretized candidate pose values for either orientation or scale, represents a distribution over the pose values.   
In contrast to previous regression-based methods~\cite{yi2016learning, ono2018lf, mishkin2018repeatability, shen2019rf}, which predict only a single pose, our histogram estimator is able to naturally predict multiple plausible poses by a multi-modal histogram distribution, and can be effectively trained with our self-supervised objective.
For scale estimation, inspired by the scale space of SIFT~\cite{lowe2004distinctive}, we create 13 bins over the $\log_2$ scale space, $\ie, B_\mathrm{s}= 13$, which are centered on $\{-2, -\frac{5}{3}, ..., 0, ..., \frac{5}{3}, 2\}$ so that each bin covers the span of $\frac{4}{B_\mathrm{s}-1}$ in $\log_2$ scale from its center. 
For orientation estimation, we create 36 bins over $2\pi$, $\ie, B_\mathrm{o}= 36$, which are centered on $\{0, \frac{\pi}{18}, ..., \frac{35\pi}{18} \}$ so that each bin covers the span of $\frac{2\pi}{B_\mathrm{o}}$ in radians from its center. 


We train our model using a self-annotated dataset of image patch pairs that are generated by transforming images with random rescaling/rotating. 
Let us assume such a dataset of image patches $\mathcal{D} = \{ (I_n, I'_n, \Delta_n ) \}_{n=1}^N$, where $\Delta_n$ denotes the ground-truth relative pose from $I_n$ to $I'_n$. 
Note that we do not have a manually labeled pose for either of the two image patches; the relative pose difference between them is the only supervisory signal we exploit. 
To train our model using the limited self-supervision, we propose the {\em histogram alignment} loss that aligns one histogram to the other and then measures the discrepancy between the aligned histograms. 
In training, a pair of image patches, $I$ and $I'$, from the dataset $\mathcal{D}$ are fed into our model to predict pose histograms of the two patches, $h$ and $h'$, respectively.  
Figure~\ref{fig:histogram_alignment} illustrates the concept of the histogram alignment losses, which will be detailed subsequently.

\smallbreak \noindent 
\textbf{Scale.} 
We define the histogram shift operator $T^d$ ($d \in \mathbb{R}$) that takes a histogram $h$ on $\mathbb{Z}$ and translates it to the left by $d$ with a linear interpolation:  
\begin{equation}
    T^{d} h(i) = 
    \begin{cases}
        h\big( i + d  \big)                                             & \text{if $d \in \mathbb{Z}$}  \\
        \big( \lceil d \rceil - d \big) h\big( i + \lfloor d \rfloor \big)  +  \big(d - \lfloor d \rfloor \big) h\big( i + \lceil d \rceil \big)     &  \text{otherwise,}
    \end{cases}
\end{equation}
where $\lfloor \cdot \rfloor$ and $\lceil \cdot \rceil$ denotes the floor and the ceiling, respectively, and any other interpolation can replace the linear one. This enables the shifting operator $T^d$ to cover a non-integer number $d$ in general. 

Let us consider an image $I$ and its scaled image $I'$ by $\Delta_\mathrm{s}$ in $\log_2$ scale. To align their scale histogram outputs, $h_\mathrm{s}$ and $h'_\mathrm{s}$, we shift $h'_\mathrm{s}$ by $\frac{(B_\mathrm{s}-1)\Delta_\mathrm{s}}{4}$, since $\frac{4}{B_\mathrm{s}-1}$ is a single bin coverage per $\log_2$ scale. 
Given the bins of $h_\mathrm{s}$, indexed by $\{ 0, 1, ..., B_\mathrm{s}-1\}$, the set of bins $\mathcal{B}$ that shares the same scales with the shifted scale histogram $T^{\frac{(B_\mathrm{s}-1)\Delta_\mathrm{s}}{4}} h'_\mathrm{s}$ is 
\begin{equation}
    \mathcal{B}  =  
    \begin{cases}
     \{ i \mid 0 \leq i \leq B_\mathrm{s} - 1 - \lceil \frac{(B_\mathrm{s}-1)\Delta_\mathrm{s}}{4} \rfloor ) \} & \text{if $\Delta_\mathrm{s} \geq 0$} \\
     
     \{ i \mid - \lceil \frac{(B_\mathrm{s}-1)\Delta_\mathrm{s}}{4} \rfloor \leq i \leq B_\mathrm{s} - 1 \} & \text{if $\Delta_\mathrm{s} < 0$}, 
   
    \end{cases}
\end{equation}
where $\lceil \cdot \rfloor$ denotes the rounding to the nearest integer. 

Finally, given the ground-truth scale shift $\Delta_\mathrm{s}$ from $I$ to $I'$, 
the histogram alignment loss for scale computes the distance between the shared parts of the two scale histograms aligned by the histogram shift:  
\begin{equation}
    \mathcal{L}_\mathrm{s}(h_\mathrm{s}, h'_\mathrm{s}) = -\sum_{i \in \mathcal{B} } h_\mathrm{s}(i)  \log \big( T^{\frac{(B_\mathrm{s}-1)\Delta_\mathrm{s}}{4} } h'_\mathrm{s}( i ) \big), 
\end{equation}
where only the bins of shared scales contribute to the loss. We use the cross-entropy to enforce the two histograms to match.

\smallbreak \noindent 
\textbf{Orientation.} 
To handle the circular property of orientation, we define the circular shift operator $T^{d}_{B}$ on $\{ 0, 1, ..., B-1 \}$: 
\begin{align}
    T_B^{d} h(i) & =
    \begin{cases}
        h\big( ( i + d ) \bmod B  \big)                   & \hspace{-0.2cm} \text{if $d \in \mathbb{Z}$} \\
        \big( \lceil d \rceil - d \big) h\big( ( i + \lfloor d \rfloor ) \bmod B \big) + \big(d - \lfloor d \rfloor \big) h\big( ( i + \lceil d \rceil ) \bmod B \big) & \hspace{-0.2cm} \text{otherwise,} 
     \end{cases}    
\end{align}
where the modulo operation uses the floored division so that the output is a non-negative integer. Note that this histogram shift can cover any rotation value of $d \in \mathbb{R}$.

Let us consider an image $I$ and its rotated image $I'$ by $\Delta_\mathrm{o}$ in radians. To match their orientation histogram outputs, $h_\mathrm{o}$ and $h'_\mathrm{o}$, we circular-shift $h'_\mathrm{o}$ by $\frac{B_\mathrm{o}\Delta_\mathrm{o}}{2\pi}$, since $\frac{2\pi}{B_\mathrm{o}}$ is a single bin coverage per radian.  
As the result, the bins of each histogram, indexed by $\{ 0, 1, ..., B_\mathrm{o} - 1 \}$, are aligned to have the same orientation. 
Therefore, given the ground-truth orientation shift $\Delta_\mathrm{o}$ from $I$ to $I'$, 
the histogram alignment loss for orientation computes the distance between the two orientation histograms aligned by the circular shift: 
\begin{equation}
    \mathcal{L}_\mathrm{o}(h_\mathrm{o}, h'_\mathrm{o}) = -\sum_{i=0}^{B_\mathrm{o}-1} h_\mathrm{o}(i)  \log \big( T_{B_\mathrm{o}}^{\frac{B_\mathrm{o}\Delta_\mathrm{o}}{2\pi} } h'_\mathrm{o}( i ) \big). 
\end{equation}

These two losses allow us to train the scale and orientation estimators without characteristic scale and orientation annotations,  
by defining the characteristic scale and orientation of an image patch in a relative manner which are consistently estimated with the other corresponding patch.
The overall training objectives for scale and orientation estimation are  
\begin{equation}
     \mathcal{L}_\mathrm{s} = \mathcal{L}_\mathrm{s}(h_\mathrm{s}, h'_\mathrm{s}) + \mathcal{L}_\mathrm{s}(h'_\mathrm{s}, h_\mathrm{s}), \,\,\,\,\,\,
      \mathcal{L}_\mathrm{o} = \mathcal{L}_\mathrm{o}(h_\mathrm{o}, h'_\mathrm{o}) + \mathcal{L}_\mathrm{o}(h'_\mathrm{o}, h_\mathrm{o}),
\end{equation}
where we use an additional term to make the objectives symmetric for the two histograms.

\begin{wrapfigure}{R}{0.4\textwidth}
    \centering
    \vspace{-0.5cm}
    \scalebox{0.25}{
    \includegraphics{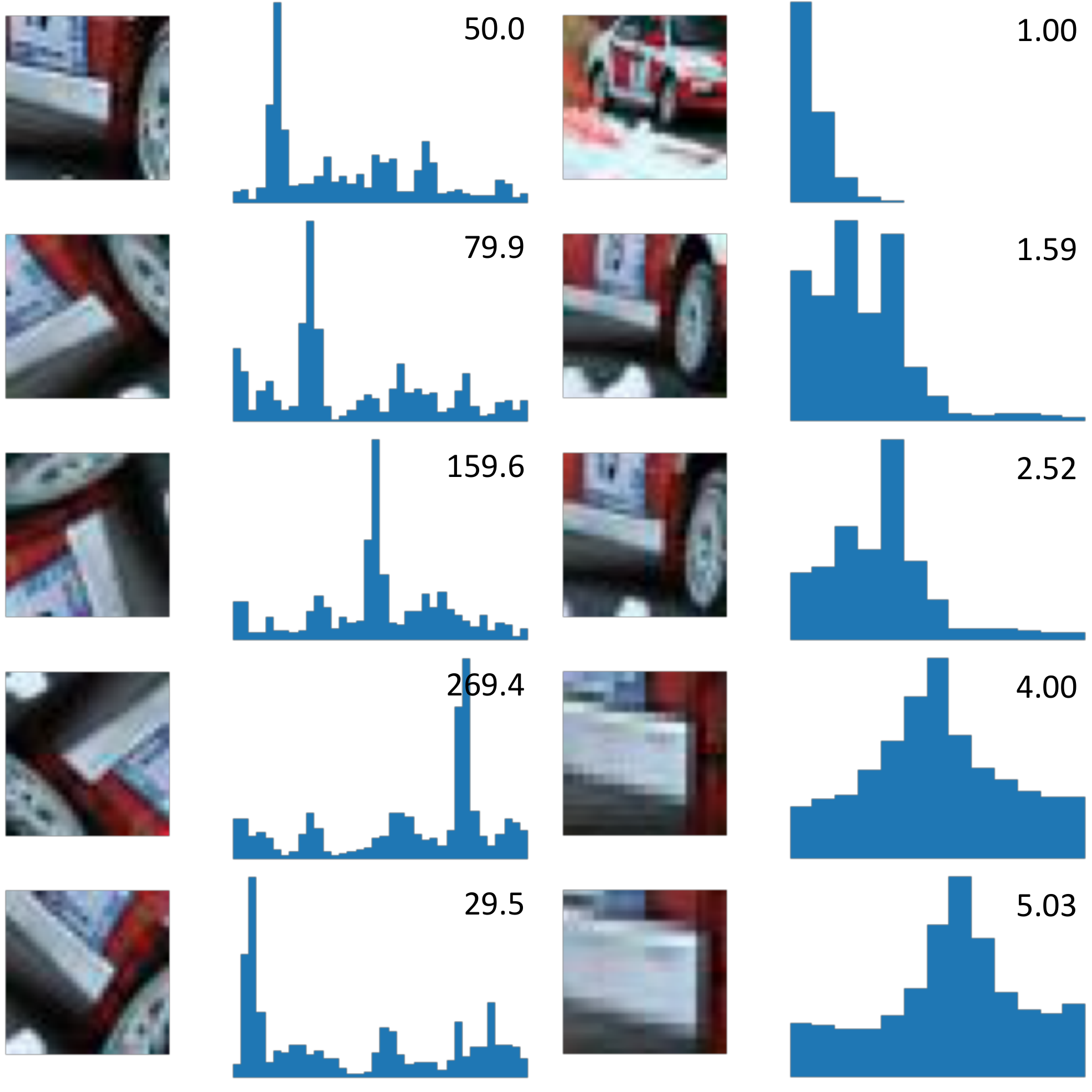}
    }
     \vspace{-0.3cm}
    \caption{Predicted rotation/scale histograms for different inputs. The numbers denote the estimated pose values. For better visualization, we suppress non-maximal values in the histograms via softmax. 
    }
    \label{fig:vis_histogram}
\end{wrapfigure}

\section{Experiments}

We conduct experiments to demonstrate the efficacy of our method. In this section, 
we explain the implementation details, describe datasets with their evaluation metrics, and then show 
experimental results with in-depth analyses. 

\subsection{Implementation details}\label{sec:implementation}
We use the ResNet-18~\cite{he2016deep} backbone as a feature extractor and train the whole networks from random initialization. 
We use two separate models for scale and orientation estimators.
Both the estimators are implemented with four-layer MLPs.
We resize the input patch size $I \in \mathbb{R}^{3 \times 32 \times 32}$.
Our model yields an orientation vector $h_\textrm{o} \in \mathbb{R}^{36}$ and a scale vector $h_\textrm{s} \in \mathbb{R}^{13}$ as outputs.
We use a batch size 64, a SGD optimizer with a learning rate 3.0 and a momentum 0.9.
We set the softmax temperature value to 20 for stable learning.

\noindent\textbf{Inference.}
We use a simple $\argmax$ function to convert the histogram to a single pose value.
\begin{equation}
    f_\textrm{s}(I) = 2^{\frac{4}{B_\mathrm{s}-1}\argmax_{i}(h_\textrm{s}(i))},\ \ 
    f_\mathrm{o}(I) = \frac{2\pi}{B_\mathrm{o}}\argmax_{i}(h_\textrm{o}(i)),
\end{equation}
where $B_\mathrm{s}$ and $B_\mathrm{o}$ are the numbers of bins for scale and orientation, respectively. 
Figure~\ref{fig:vis_histogram} visualizes predicted orientation/scale histograms for different input patches. 
While further non-maximum suppression or smoothing schemes can also be adopted for the histograms~\cite{lee2019sfnet}, 
we use the simple $\argmax$ function to determine the final pose in our work.

\subsection{Evaluation Benchmarks}\label{sec:benchmarks}
We use three datasets, PatchPose, HPatches~\cite{balntas2017hpatches} and IMC2021~\cite{jin2021image}. The PatchPose dataset is constructed by us for learning and evaluation; our model is trained on its train split and tested on its test split. HPatches~\cite{balntas2017hpatches} and IMC2021~\cite{jin2021image} are employed to evaluate the transferability of the learned model; they are used for evaluation only. 

\noindent
\textbf{PatchPose dataset generation.}~\label{sec:data_generation}
The PatchPose dataset is synthetically generated from 1,793 images of SPair-71k~\cite{min2019spair} from PASCAL-VOC~\cite{everingham2010pascal}. 
We extract 3 keypoints of an image using SIFT~\cite{lowe2004distinctive}, on which $64 \times 64$ patches are centered on to be cropped after transformed by $\Delta_\textrm{s}, \Delta_\textrm{o} \in \mathbb{R}$.  
We pair the source patch and its augmented patch for inputs to the network.
The dataset of patch pairs with relative pose annotation
is generated without manual annotation. 
The used rescaling and rotating degrees, $\Delta_\textrm{s}$ and $\Delta_\textrm{o}$, are annotated for free. 
The values for rescaling $\Delta_\textrm{s}$ are distributed in the range of $[2^{-2}, 2^2]$  and those for rotating $\Delta_\textrm{o}$ are in the range of $[0, 2\pi)$, covering wide ranges of scale and orientation changes. 
The dataset is split as train : val : test $=$ 3,947,054 : 40,276 : 40,278. 
For details, see the supplementary material.

\noindent\textbf{Evaluation metric.}
To evaluate predicted poses of two image patches, we define accuracy metrics. 
We first measure the errors using $\log_2$-scale and radian-orientation differences:  
\vspace{-0.1cm}
\begin{equation}\vspace{-0.3cm}
    s(I, I'; f_\textrm{s}, \Delta_\textrm{s}) = |\log_2(\frac{f_\textrm{s}(I')}{f_\textrm{s}(I)}) - \Delta_\textrm{s}|, \ \
    o(I, I'; f_\textrm{o}, \Delta_\textrm{o}) = |(f_\textrm{o}(I') - f_\textrm{o}(I)) \bmod 2\pi - \Delta_\textrm{o}| ,
\end{equation}
where $I$ and $I'$ are the image pair with known difference in scale $\Delta_\textrm{s}$ and that in orientation $\Delta_\textrm{o}$. $f_\textrm{s}$ and $f_\textrm{o}$ are scale and orientation estimators, respectively. 
We then convert the errors to accuracy values using some thresholds, \ie, \{$\frac{1}{6}, \frac{1}{3}$\} for scale and \{$\frac{\pi}{36}$, $\frac{\pi}{18}$\} for orientation.

\noindent
\textbf{Transferability evaluation.}
We also use the HPatches~\cite{balntas2017hpatches} viewpoint variation for transferability evaluation.
The HPatches viewpoint variation has 59 scenes; each scene has 6 images with known homography matrices. 
The ground-truth scale and orientation are extracted from homography $A_{3\times3}$ by 
\vspace{-0.4cm}
\begin{equation}\vspace{-0.1cm}
    \Delta_\textrm{s} = \sqrt{(\frac{A_{11}}{A_{33}})^2 + (\frac{A_{21}}{A_{33}})^2 },  \ \ \ 
    \Delta_\textrm{o} = \arctan(\frac{A_{21}}{A_{11}}). \\
\end{equation}
We extract 25 patches centered on SIFT~\cite{lowe2004distinctive} and Harris~\cite{Harris88acombined} keypoints, and then extract patches centered on the corresponding keypoints from the other image.
As the result, we sample 7,375 patch pairs used for the pose estimation evaluation on HPatches~\cite{balntas2017hpatches}. 
On the other hand, we use the all 116 sequences (59 viewpoint, 57 illumination) of  HPatches~\cite{balntas2017hpatches} to evaluate our method on image matching.
HPatches is an image matching benchmark with ground-truth homography.  
We evaluate our patch extraction ability on image matching pipeline compared to the existing methods~\cite{lowe2004distinctive, mishchuk2017working, tian2019sosnet}.
For each sequence, we pair the first image to 5 other images, so a total of 580 image pairs are used.
To evaluate patch extraction on the image matching, we use the number of matches and mean matching accuracy (MMA) as evaluation metrics. 

To demonstrate the effectiveness of our method on a more complex dataset, we use the IMC2021~\cite{jin2021image} wide-baseline matching benchmark. IMC2021~\cite{jin2021image} consists of an unconstrained urban scene with large illumination and viewpoint variations; the validation set of Phototourism and Pragueparks are used to evaluate our method. This benchmark takes matches as input and measures the quality of 6 DoF pose estimation. We use our predicted patch scale/orientation for the outlier rejection~\cite{cavalli2020handcrafted} scheme in the image matching pipeline~\cite{lowe2004distinctive, barroso2019key, mishchuk2017working, cavalli2020handcrafted, chum2005two} and measure the mean average accuracy (mAA) at 5\degree and 10\degree of the pose estimation and the number of inliers as evaluation metrics.
\begin{table}[t]
\centering
\begin{tabular}{c||cc|cc||cc|cc}
\multirow{3}{*}{methods~} & \multicolumn{4}{c||}{PatchPose  } &  \multicolumn{4}{c}{HPatches  } \\ \cline{2-9}
& \multicolumn{2}{c|}{sca. ($\mathrm{log}_2$) }   & \multicolumn{2}{c||}{ori. ($\mathrm{radian}$) }    & \multicolumn{2}{c|}{sca. ($\mathrm{log}_2$) }   & \multicolumn{2}{c}{ori. ($\mathrm{radian}$) }    \\ 
\cline{2-9}
 & $\pm\frac{1}{6}$ & $\pm\frac{1}{3}$      & $\pm \frac{\pi}{36}$ & $\pm \frac{\pi}{18}$  & $\pm\frac{1}{6}$ & $\pm\frac{1}{3}$      & $\pm \frac{\pi}{36}$ & $\pm \frac{\pi}{18}$        \\   \hline
SIFT~\cite{lowe2004distinctive}             &  \underline{28.3}  & \underline{44.9}  & 15.5   &  28.7    &  \underline{11.3}  & 25.3   & 11.2    & 25.6  \\
OriNet~\cite{yi2016learning}                &  -  & -   &  \underline{29.1}  & \underline{45.0}   &  -  &  -  &  \underline{15.8}  & 29.8  \\
LF-Net~\cite{ono2018lf}                     & 10.6   & 17.6   & 13.7   & 25.3  & 8.1    & 25.5   & 14.2   & 24.5    \\
AffNet~\cite{mishkin2018repeatability}      & -   & -   &  27.0  &  42.0  &  -    &  -   & 14.0   & 23.9  \\
RF-Net~\cite{shen2019rf}                    & 10.6   & 17.4   & 4.0  & 6.6    & 7.8   & \underline{26.1}   & 15.6    & \underline{32.9}  \\
\hline
ours         &  \textbf{57.9}  & \textbf{78.2}   & \textbf{80.5}   & \textbf{97.9}  & \textbf{29.0}   & \textbf{53.0}   & \textbf{52.0}   & \textbf{69.2}  \\
\end{tabular}
\caption{Accuracy of patch pose estimation on the PatchPose and the HPatches viewpoint variation.  
The bold numbers indicate the best and the underlined ones are the second best.\protect\footnotemark[1]
}\label{tab:scale_estimation}
\end{table}
\footnotetext[1]{  We measure the scores of OriNet~\cite{yi2016learning}, AffNet~\cite{mishkin2018repeatability}, LF-Net~\cite{ono2018lf} and RF-Net~\cite{shen2019rf} using official released code by authors. SIFT~\cite{lowe2004distinctive} score is measured by modification of OpenCV model. 
}
\vspace{-0.2cm}
\subsection{Patch Pose Estimation}\label{sec:results}
\vspace{-0.2cm}
We evaluate our method, which is trained using the PatchPose training split, and compare it with the other methods~\cite{lowe2004distinctive, yi2016learning, ono2018lf, mishkin2018repeatability, shen2019rf} on the PatchPose test split and the HPatches~\cite{balntas2017hpatches} viewpoint variation.

\noindent \textbf{PatchPose.}
The left side of Table~\ref{tab:scale_estimation} shows patch pose estimation results on the PatchPose test split compared to the existing methods~\cite{lowe2004distinctive, yi2016learning, ono2018lf, mishkin2018repeatability, shen2019rf}. 
Our method outperforms the previous methods by a large margin at all thresholds in both scale/orientation estimation. 
In particular, our orientation estimator achieves an almost perfect accuracy of 97.9\% at $\frac{\pi}{18}$ threshold. 
The regression-based learning methods~\cite{yi2016learning, ono2018lf, mishkin2018repeatability, shen2019rf}, which learn to estimate scale/orientation implicitly by improving descriptor similarity, turn out to perform significantly worse than ours. 


\noindent \textbf{HPatches.}
The right side of Table~\ref{tab:scale_estimation} shows the results on the HPatches~\cite{balntas2017hpatches} viewpoint variation. 
 The orientation estimation results of RF-Net~\cite{shen2019rf} on HPatches~\cite{balntas2017hpatches} performs better than those on PatchPose; we find this is because (1) RF-Net is trained on the subset of HPatches and (2) the limited range of RF-Net orientation prediction coincides more with the true orientation range of HPatches.
Note that the other methods~\cite{yi2016learning, ono2018lf, mishkin2018repeatability}, including ours, have not been trained on this HPatches dataset.
These results show that our self-supervised model transfers well to unseen patches from a different domain with unseen transformations. 
For all the methods, the scores on HPatches are lower than those on PatchPose due to the shear and/or tilt factors of transformation in image pairs from HPatches, which renders scale/orientation prediction more challenging.

\begin{wraptable}{R}{0.45\textwidth}
\centering
\scalebox{0.9}{
\begin{tabular}{c|l||cc|cc}
\multirow{2}{*}{} & \multirow{2}{*}{top-$k$}     &   \multicolumn{2}{c}{sca. ($\mathrm{log}_2$)}             & \multicolumn{2}{|c}{ori. ($\mathrm{radian}$)}    \\ \cline{3-6}
& &  $\pm\frac{1}{6}$ & $\pm\frac{1}{3}$      & $\pm \frac{\pi}{36}$ & $\pm \frac{\pi}{18}$                         \\
\hline
\multirow{4}{*}{\rotatebox[origin=c]{90}{SIFT}}
& top-1 &      29.3 & 44.9  & 15.5  & 28.7                   \\
& top-2 &      55.4 & 65.2  & 29.4  & 44.3                  \\
& top-3 &      68.6 & 74.8  & 41.6  & 57.1                 \\
& top-4 &      78.0 & 84.7  & 59.3  & 75.6                  \\ \hline
\multirow{4}{*}{\rotatebox[origin=c]{90}{ours}}
& top-1 &       57.9    & 78.2    & 80.5    &  97.9                   \\
& top-2 &       76.4    & 84.8    & 99.0    &  99.4                   \\
& top-3 &       83.4    & 88.8    & 99.8    &  99.9                   \\
& top-4 &       \textbf{87.8}    & \textbf{93.4}    & \textbf{99.9}    &  \textbf{100.0}                  \\
\end{tabular}
}  
\caption{Recall of histogram-based methods on PatchPose using top-$k$ candidates. }
\label{tab:topk_evaluation}
\end{wraptable}

\noindent\textbf{Multi-pose estimation.}\label{sec:topk_eval}
Unlike regression-based methods~\cite{yi2016learning, ono2018lf, mishkin2018repeatability, shen2019rf}, the histogram-based methods, ours and SIFT~\cite{lowe2004distinctive}, can naturally leverage multiple candidates of scale and orientation for each image patch by selecting multiple modes from the predicted histograms. 
To observe the potential gain of using multiple candidates, we select the top-$k$ scale/orientation candidates for each image patch and measure whether a true pair of scale/orientation predictions is present between two corresponding sets of the top-$k$ candidates. 
Table~\ref{tab:topk_evaluation} shows the recall performance on the PatchPose test split, where we vary the number of candidates from 1 to 4.  
The results show that our method substantially outperforms the classical histogram-based method, SIFT~\cite{lowe2004distinctive}.
In particular, our model achieves 100\% recall of orientation estimation in top-4 selection at $\frac{\pi}{18}$ threshold.
The use of multiple candidates allows effective image matching in Sec.~\ref{sec:image_matching}, which is not available for regression-based methods.

\begin{wraptable}{R}{0.45\textwidth}
\centering
\vspace{-0.2cm}
\scalebox{0.9}{
\begin{tabular}{l|cc||l|cc}
 \multirow{2}{*}{$B_\mathrm{s}$}     &   \multicolumn{2}{c||}{sca. ($\mathrm{log}_2$)}             & \multirow{2}{*}{$B_\mathrm{o}$} & \multicolumn{2}{|c}{ori. ($\mathrm{radian}$)}    \\ \cline{2-3}\cline{5-6}
&     $\pm\frac{1}{6}$ & $\pm\frac{1}{3}$   &   & $\pm \frac{\pi}{36}$ & $\pm \frac{\pi}{18}$                      \\
\hline
7 & 22.5 &  22.5     & 9   & 26.5  & 26.5                   \\
9 & 37.3 & 37.3      & 18  & 48.7  & 48.7                  \\
13 & \textbf{57.9} & \textbf{78.2}     & 36  & \textbf{80.5}  & \textbf{97.9}                 \\
17 & 19.7 & 35.6     & 72  & 9.7   & 15.4                  \\ 
\end{tabular}
}  
\caption{Accuracy of patch pose estimation on PatchPose with different numbers of bins $B$.}
\label{tab:design_choice}
\end{wraptable}

\noindent \textbf{Effect of histogram sizes.} 
Table~\ref{tab:design_choice} shows the accuracy variations with different numbers of bins $B_\mathrm{s}$ and $B_\mathrm{o}$. As the number increases to a proper value, the histogram becomes more fined-grained and thus the prediction tends to be more precise. 
However, when it increases further (\eg, $B_\mathrm{s}=17, B_\mathrm{o}=72$), we find the training process becomes unstable due to the increased classes for prediction. We thus set the values as $B_\mathrm{s}=13$ and  $B_\mathrm{o}=36$.

\begin{figure}[t]
    \begin{minipage}{0.65\linewidth}
    \scalebox{1.05}{
        \includegraphics[width=78mm]{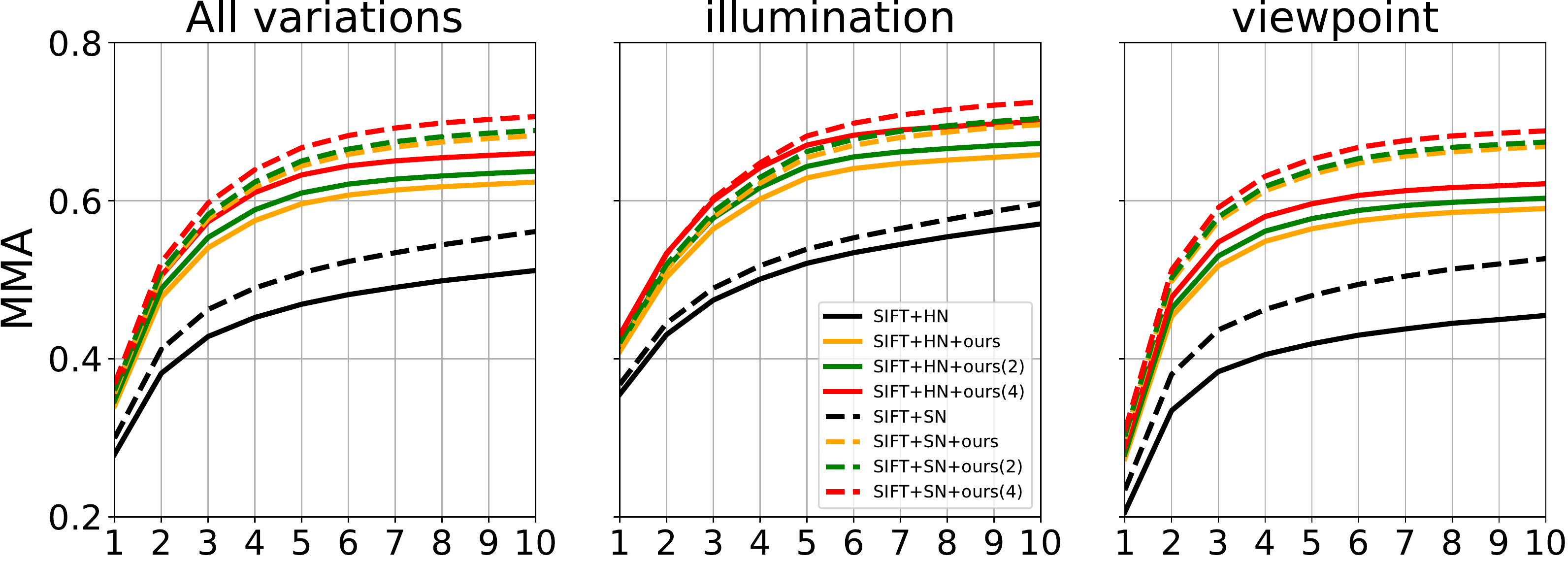}
        }
    \end{minipage} \hspace{-0.5cm} \hfill 
    \begin{minipage}{0.34\linewidth}
        \scalebox{0.65}{
        \begin{tabular}{@{} c|c||c|c @{}}
        \multicolumn{2}{c||}{methods}       & \multirow{2}{*}{M}    &  \multicolumn{1}{c}{MMA} \\ 
        \cline{1-2}\cline{4-4}
        Det. + Des. & Pose.   &     & 3px                                              \\  
        \hline
        \multirow{4}{*}{ \specialcell{SIFT+HN \\ ~\cite{lowe2004distinctive, mishchuk2017working}} } & SIFT         &    148.9    &  0.45                  \\
            & ours                                     &    154.6    &  0.57           \\ 
            & ours (top2)                              &    153.8    &  0.59                \\ 
           &  ours (top4)                              &    \textbf{157.6}    &  \textbf{0.61}                   \\ \hline
        \multirow{4}{*}{ \specialcell{SIFT+SN \\ ~\cite{lowe2004distinctive, tian2019sosnet}} } & SIFT            &    144.3    &  0.49                \\
            & ours                              &    158.8    &  0.62                   \\ 
            & ours (top2)                      &    157.7    &  0.62                    \\ 
            & ours (top4)                      &    \textbf{160.1}   &  \textbf{0.64}                    \\  \hline%
        \end{tabular}
        }  
        \end{minipage}
        \caption{Mean matching accuracy (\%) with off-the-shelf keypoint detectors and descriptors on HPatches. 
        The number beside ours means the number of top-$k$ candidates.
        `M' denotes the average number of matches. 
        We fix the average number of keypoints all the same.
        }\label{tab:matching}
        
\end{figure}
\vspace{-0.2cm}
\subsection{Application to Image Matching}\label{sec:image_matching} We validate our patch pose estimators by applying them to image matching. In the matching pipeline, keypoints and their scale/orientation pose values are extracted from images by an existing detector. Basically, we replace their pose values with our results for comparison.

\noindent \textbf{Evaluation on HPatches.}
In this matching accuracy evaluation,  two sets of image patches are extracted from an image pair using detected keypoints and their estimated patch poses\footnote[2]{To avoid sampling patches from outside of the image, we exclude keypoints near boundaries, \ie $(w < 16) \lor (h < 16) \lor (w > W-16) \lor (h > H-16)$ where $(W, H)$ denotes the image size and $(w, h)$ is the keypoint coordinate.} and then are matched via mutual nearest neighbors according to the similarity of patch descriptors;  SIFT~\cite{lowe2004distinctive} is used for keypoint detection while  HardNet~\cite{mishchuk2017working} and SOSNet~\cite{tian2019sosnet} are for patch description. 
In this matching pipleline, we use our pose estimation for image patch extraction and evaluate its effect. 
To leverage our multi-pose estimation in matching, we extract multiple patches for each keypoint using its top-$k$ poses. 
Figure~\ref{tab:matching} shows the image matching results on HPatches~\cite{balntas2017hpatches}, 
where the use of our method for patch extraction consistently improves over the baseline methods.
Even without multi-pose estimation, our method achieves better Mean Matching Accuracy (MMA) than all of the baselines.
Our result with the top-4 pose estimation improves both MMA and the number of matches. 
It shows that our method transfers well to image matching without any fine-tuning on the target datasets.

\begin{table}[t]
\centering
\scalebox{0.8}{
\begin{tabular}{l|c||ccc|ccc}
 \multirow{2}{*}{Det.+Pose.} & \multirow{2}{*}{K}     &   \multicolumn{3}{c|}{Phototourism} &  \multicolumn{3}{c}{Pragueparks}              \\ \cline{3-5} \cline{6-8}
&   &  Num. Inl. &  mAA(5\degree)   & mAA(10\degree) &  Num. Inl. &       mAA(5\degree)   & mAA(10\degree)              \\\hline
SIFT+AffNet~\cite{lowe2004distinctive, mishkin2018repeatability}    & 1,024           &  46.4    & 0.250 & 0.321 & 35.9 & 0.090 & 0.145 \\ 
SIFT+ours      & 1,024           &  \textbf{60.8}    & \textbf{0.316} & \textbf{0.397} & \textbf{51.3} & \textbf{0.197} & \textbf{0.277}   \\ \hline
SIFT+AffNet~\cite{lowe2004distinctive, mishkin2018repeatability}    & 2,048           &  110.9   & 0.448 & 0.542 & 90.7 & 0.196 & 0.282               \\
SIFT+ours      & 2,048           &  \textbf{131.7} & \textbf{0.471} & \textbf{0.566} & \textbf{112.8} & \textbf{0.291} & \textbf{0.401}                \\ \hline
Key.Net~\cite{barroso2019key}    & 1,024           &  75.4 & 0.319 & 0.409 & \textbf{175.9} & 0.422 & 0.562             \\
Key.Net+ours      & 1,024           & \textbf{77.6} & \textbf{0.329} & \textbf{0.420} & 175.4 & \textbf{0.443} & \textbf{0.583}               \\ \hline
Key.Net~\cite{barroso2019key}    & 2,048           &  167.5 &	0.431	& 0.537	& \textbf{368.8}	& 0.514 &	\textbf{0.660}             \\
Key.Net+ours      & 2,048           &  \textbf{172.4} & \textbf{0.446} & \textbf{0.553} & 368.7 & \textbf{0.518} & \textbf{0.660}              \\ 
\end{tabular}
}  
\caption{Mean average accuracy (mAA; 5\degree, 10\degree) of 6-DoF pose estimation and the number of inlier matches (Num. Inl.) on IMC2021~\cite{jin2021image} validation set. } \vspace{-0.3cm}
\label{tab:imc_2021}
\end{table}

\noindent \textbf{Evaluation on IMC2021.} 
In this 6 DoF camera pose estimation evaluation, we collect reliable feature matches and use them to estimate a camera pose\footnote[3]{ We evaluate to use the provided source code from IMC2021. \href{https://github.com/ubc-vision/image-matching-benchmark}{https://github.com/ubc-vision/image-matching-benchmark}} via the standard structure-from-motion method~\cite{schonberger2016structure}. To obtain the set of reliable matches, we first obtain mutual nearest neighbor matches via a standard feature extraction and matching process~\cite{lowe2004distinctive, mishkin2018repeatability, mishchuk2017working}, and then purify those matches using the outlier rejection method of AdaLAM~\cite{cavalli2020handcrafted} and the robust model fitting of DEGENSAC~\cite{chum2005two}. 
In this pipeline, we use our estimated patch poses for the outlier rejection step of AdaLAM~\cite{cavalli2020handcrafted}. 
For comparison, we use SIFT+AffNet~\cite{lowe2004distinctive, mishkin2018repeatability} and Key.Net~\cite{barroso2019key} as two baselines for keypoint detection and patch pose estimation, and evaluate the effect of replacing their orientation estimation with ours in the process of outlier rejection. 
For all the methods, HardNet~\cite{mishchuk2017working} is used for patch description. 
 Table~\ref{tab:imc_2021} shows the results of 6 DoF pose estimation in the validation set of the IMC2021 stereo task~\cite{jin2021image}. 
Our method improves over SIFT+AffNet~\cite{lowe2004distinctive, mishkin2018repeatability} and Key.Net~\cite{barroso2019key} in 6 DoF pose estimation accuracy on both Phototourism and Pragueparks. 
The performance gain of our method on Key.Net is smaller than that on SIFT+AffNet~\cite{lowe2004distinctive, mishkin2018repeatability}. We find this is due to the keypoint selection scheme of Key.Net~\cite{barroso2019key}; it selects the keypoints based on the local window so that they spread evenly, which reduces the impact of the subsequent outlier rejection step. 

\section{Conclusion}
We have proposed a self-supervised learning framework for characteristic scale and orientation estimation. 
Our method effectively estimates characteristic scale and orientation via the histogram alignment technique. 
Our experiments show impressive results on PatchPose and HPatches datasets, achieving the state-of-the-art performance on the task of scale and orientation estimation.
Moreover, the use of our patch pose estimation has been shown to improve matching performance on HPatches and IMC2021 benchmarks, on which our method has never been trained. 
We believe further research in this direction can benefit a variety of image matching, visual localization, and recognition problems in computer vision. 


\section*{Acknowledgement}
This work was supported by Samsung Electronics Co., Ltd., the NRF grant (NRF-2021R1A2 C3012728; NRF-2019H1A2A1076171 - Global Ph.D. Fellowship Program), and the IITP grant (No.2019-0-01906, AI Graduate School Program - POSTECH) funded by Ministry of Science and ICT, Korea.

\bibliography{egbib}
\end{document}


\maketitle
\vspace{-0.5cm}

This supplementary material consists of four parts. 
Section~\ref{sec:supp_data_generation} describes a more detailed process of our PatchPose dataset generation.
Section~\ref{sec:supp_matching} shows our scale and orientation estimation results on the image matching tasks and in-depth analysis. 
Section~\ref{sec:supp_pred_range} analyzes the prediction range of scale and orientation compared to the existing methods. 
Section~\ref{sec:supp_qual} shows various qualitative results to verify our model.
\begin{wrapfigure}{R}{0.5\textwidth}
    \centering
    \scalebox{0.5}{
    \includegraphics{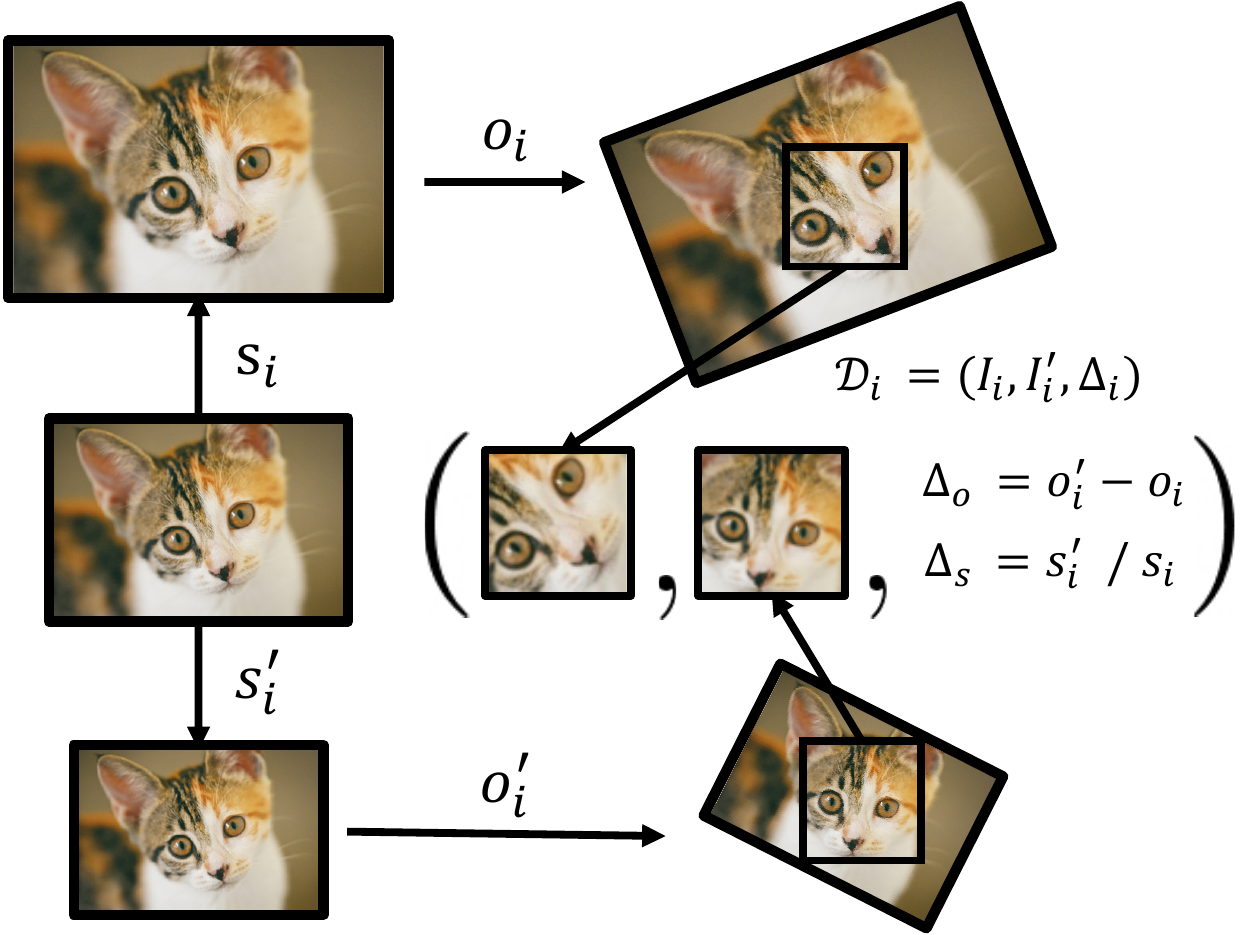}
    }
     \vspace{-0.2cm}
    \caption{Illustration of training data generation. 
    Each entry of $\mathcal{D}$ stores two cropped patches extracted about the same keypoint from differently augmented versions of an image and their ground-truth scale/orientation differences. 
    }
    \label{fig:example_training_data}
\end{wrapfigure}

\section{PatchPose dataset generation}\label{sec:supp_data_generation}
We generate our PatchPose dataset, extracted from 1,793 images of SPair-71k~\cite{min2019spair} from PASCAL-VOC~\cite{everingham2010pascal}.
The PatchPose dataset consists of two parts, PatchPose-A and PatchPose-B.
Figure~\ref{fig:example_training_data} shows to generate an example of the PatchPose, $\mathcal{D} = \{ (I_n, I'_n, \Delta_n ) \}_{n=1}^N$, where $\Delta_n$ denotes the ground-truth relative pose from $I_n$ to $I'_n$.

\noindent \textbf{Dataset specification.}
The PatchPose-A dataset contains all combinations of scale and orientation shifts over 36 rotation and 13 scale values. 
We first produce 2,517,372 patches and then prune them as described in the next subsection of this supplementary material. 
After pruning, we obtain 2,013,804 patches, which make pairs with their original patches from the original images.
We split the PatchPose-A dataset into 1,973,527 training pairs, 20,138 validation pairs, and 20,139 test pairs, whose split ratio is train : val: test = $98:1:1$.
The PatchPose-A dataset has $\frac{1}{3}$ interval of $\log_2$-scale in the range of $[-2, 2]$ and $\frac{1}{18}\pi$ interval of orientation in the range of $[0, 2\pi)$, covering wide ranges of scale and orientation changes.
In the PatchPose-A dataset, the orientation value is fixed when the scale value varies and vice versa. 
The PatchPose-B dataset is designed to evaluate the robustness under simultaneous and continuous changes of scale and orientation. 
Unlike the PatchPose-A, the scale and orientation of an image patch are simultaneously transformed by random value in $[\frac{1}{4}, 4]$ and $[0, 2\pi)$. 
The PatchPose-B contains the same number of patches as the PatchPose-A, which are randomly transformed by $\Delta_s, \Delta_o \in \mathbb{R}$. 
Finally, we make the PatchPose dataset to merge PatchPose-A and -B, which dataset split is train: val: test $=$ 3,947,054 : 40,276 : 40,278.

\noindent \textbf{Dataset pruning.}
Among the generated image patches, there are ambiguous patches that do not have distinct patterns for characteristic scale and orientation.
We find that those ambiguous samples often distract the learning process.

\begin{wrapfigure}{R}{0.5\textwidth}
\centering
\scalebox{0.5}{
\includegraphics{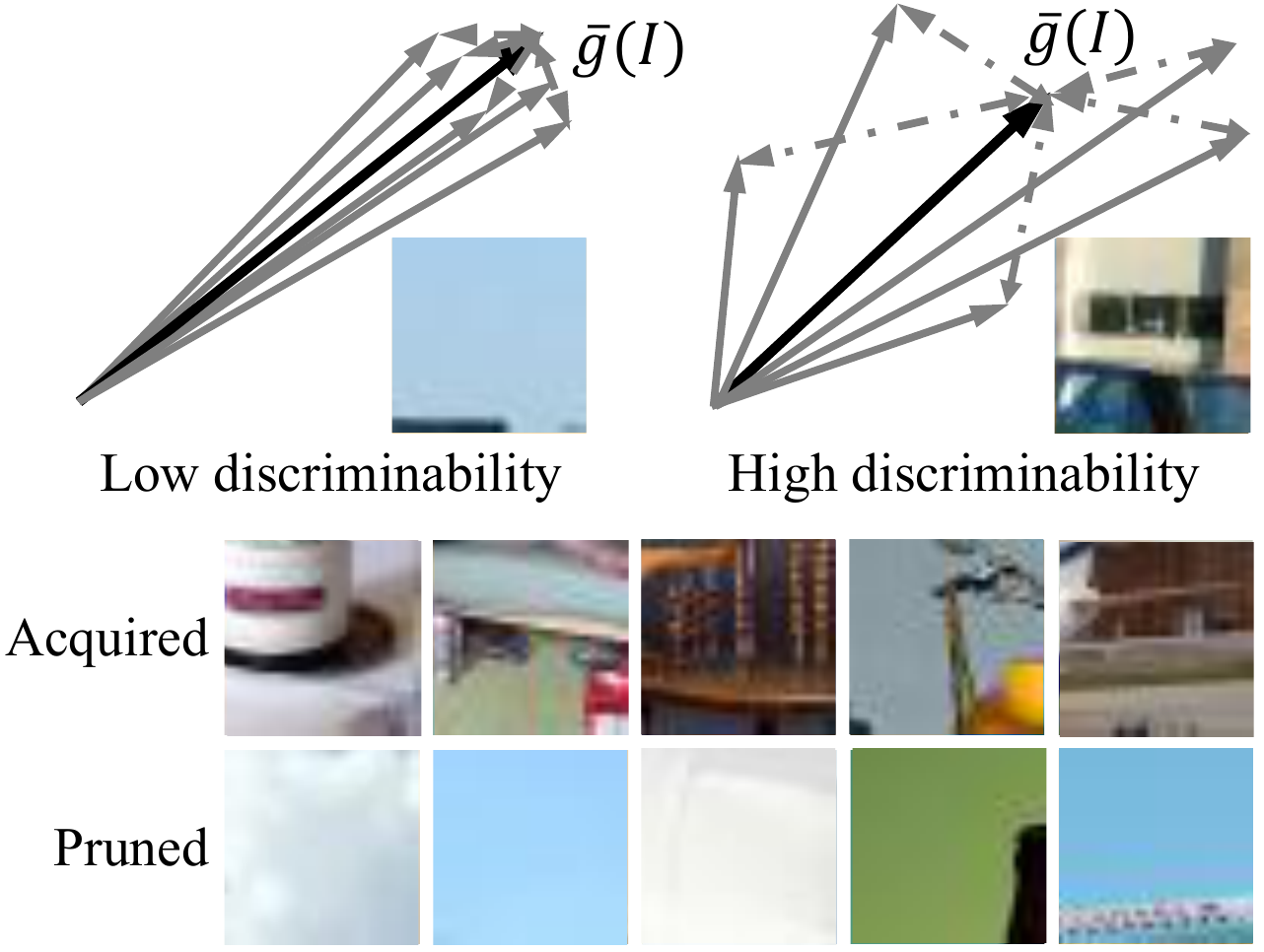}
}
\caption{Illustration of the instance discrimination scheme. 
    The left side of the top illustrates an example of a pattern-less patch with low discriminability. The source vectors of transformed patches have fewer separations from their mean vector $\bar{g}$. 
    The right side of the top illustrates a patch with high discriminability. The source vectors are sufficiently separated from the mean vector $\bar{g}$.
}
\label{fig:discriminate_vector}
\end{wrapfigure}

In order to prune the scale- and rotation-agnostic examples, we adopt an instance discrimination scheme inspired by~\cite{feng2019self}.
We measure the average standard deviation $\sigma$ of the transformed feature vectors $g(\cdot) \in \mathbb{R}^{N}$:  
\begin{equation}
    \bar{g}(I) = \frac{1}{|\mathcal{A}|
    } \sum_{(s, o) \in \mathcal{A}} g(\mathcal{T}_{s, o}(I)),
\end{equation}
\begin{equation}
    \sigma^2(I) = \sum_{i=1}^{N} \sum_{(s, o) \in \mathcal{A}}  \big(g(\mathcal{T}_{s, o}(I))_i - \bar{g}(I)_i\big)^2 ,
\end{equation}
\noindent  where $\mathcal{A}$ is a set of all possible pairs using 13 scaling and 36 rotating, \ie, $\mathcal{A} = \{(s_i, o_i)\}_{i=1}^{468}$ and $I$ is the transformed input patches. $g$ is a function that forwards to CNN to generate features from the input image. $\bar{g}(\cdot)$ denotes the mean feature vector from the set of transformed feature vectors. 

The standard deviation $\sigma$ means how the sample can be discriminated by scale and rotation. For feature extraction of pruning, we use a ResNet-18~\cite{he2016deep} model pretrained on ImageNet~\cite{imagenet_cvpr09} to extract features before the last fully connected layer.
We aim to filter the unrelated samples with scale/orientation variation (e.g., textureless, round shape), to focus on learning with clear examples. Finally, we prune $20\%$ of patches with low discriminability.  
Figure~\ref{fig:discriminate_vector} illustrates the discriminability of feature vectors and several examples of acquired and pruned patches.
Most of the pruned patches are less discriminative about scale and rotation compared to the acquired patches. 



\section{Additional analysis on image matching}\label{sec:supp_matching}
We experiment with our scale and orientation estimation on image matching methods for in-depth analysis. 
We use the HPatches~\cite{balntas2017hpatches} benchmark, and the evaluation scheme is the same as Section 4.2 and 4.4 in the main paper.
We add more baselines~\cite{lowe2004distinctive, mishchuk2017working, ono2018lf, shen2019rf, barroso2019key} and evaluate additional thresholds and report separated results of the illumination and viewpoint variations.
To evaluate the effect of our method, we replace the scale and orientation values in the image matching pipelines~\cite{lowe2004distinctive, mishchuk2017working, ono2018lf, shen2019rf, barroso2019key} at the patch extraction stage. 
We use all the pre-trained models and source codes released by the authors.

Table~\ref{tab:matching_both} summarizes the results on the HPatches dataset. 
For keypoint detectors, we use SIFT~\cite{lowe2004distinctive} and Key.Net~\cite{barroso2019key}; we use HardNet~\cite{mishchuk2017working} for descriptor extraction. Additionally, we use detection-then-description image matching methods LF-Net~\cite{ono2018lf} and RF-Net~\cite{shen2019rf}.
In addition to Figure 4 in the main paper, we report the lower/upper bounds of patch extraction.
Column `Det.' denotes the keypoints detection methods, and `Des.' denotes descriptor extraction methods, and `Pose.' denotes patch extraction methods.
The row with `ours' denotes  the results with \texttt{argmax} selection.
The row with `ours (top-k)' denotes top-k candidates selection on our output histogram as scale and orientation values. 
To measure the lower bound, we use identity matrix to extract patches centered on the keypoints as denoted the row with `lower'.
To measure the upper bound, we use the ground-truth homography matrix to extract patches centered on the keypoints as denoted the row with `upper'. 

Our model consistently performs better than the baseline methods on the mean matching accuracy (MMA) of the overall sequences and the viewpoint variation sequences.
In the illumination variations, RF-Net~\cite{shen2019rf} and Key.Net+HardNet~\cite{barroso2019key, mishchuk2017working} show robust results on several thresholds. This is because RF-Net~\cite{shen2019rf} trains their model using the HPatches dataset, and Key.Net~\cite{barroso2019key} uses synthetic training data to consider photometric variations. 
In the viewpoint variations, our model consistently performs better than the existing methods in the same settings that all the methods are trained to consider the geometric variations at training time.

\noindent \textbf{Upper bound and lower bound.}
Row 'lower' and 'upper' denote the lower bound and upper bound accuracies of patch extraction on the image matching pipeline.
We measure the lower bound using no pose extraction at all, \ie, patch sampling by an identity matrix, and the upper bound using ground-truth pose values to evaluate the test oracle of patch extraction on image matching pipelines. 
The results imply that there is still room for improvement as the upper bound (\textit{i.e.} using ground-truth pose values for pose extraction) yields the best results, motivating further research in this aspect.

\begin{table}[t!]
\centering
\scalebox{0.8}{
\begin{tabular}{c|c|c|c||c|cc|cc|cc}
\multicolumn{3}{c|}{\multirow{2}{*}{methods}}  &  \multirow{3}{*}{K}     & \multirow{3}{*}{M}    &  \multicolumn{6}{c}{MMA} \\
\cline{6-11}
\multicolumn{3}{c|}{} & & &  \multicolumn{2}{c|}{Overall} & \multicolumn{2}{c|}{Illumination} & \multicolumn{2}{c}{Viewpoint}  \\
\cline{1-3}\cline{6-11}
Det. & Des. & Pose. &   &     & 3px                       & 6px      & 3px                       & 6px       & 3px                       & 6px                        \\  
\hline
\multirow{10}{*}{SIFT~\cite{lowe2004distinctive}} & \multirow{10}{*}{HardNet~\cite{mishchuk2017working}}  
    & lower                                    & \multirow{5}{*}{122.9}  & 35.5  & 44.9   & 46.1  & 49.9  & 51.0  & 40.1  & 41.3  \\ 
 &  & SIFT                                    &                          & 35.2  & 41.7   & 44.6  & 45.8  & 48.9  & 37.7  & 40.4                          \\
 &  & ours                                    &                          & 36.3  & 51.3   & 54.3  & 53.8  & 57.2  & 48.9  & 51.5         \\ 
 &  & ours (top-4)                              &                         & \textbf{36.7}  & \textbf{54.4}  &  \textbf{57.7}   & \textbf{57.3}  & \textbf{61.1}  & \textbf{51.7}  & \textbf{54.5}     \\ 
 &  & upper                                    &                         & 38.3  & 65.5    & 69.2     & 60.7    & 64.6    & 70.2     & 73.7    \\  \cline{3-11}
 &  & lower                                    &  \multirow{5}{*}{487.6} & 151.3   & 47.8    & 49.1    & 53.0    & 54.5    & 42.7     & 43.8                         \\
 &  & SIFT                                    &                          & 148.9   & 45.2     & 49.0    & 50.1    & 54.5    &  40.5    &    43.8                      \\
 &  & ours                                    &                          & 154.6   & 57.5   & 61.4    & 60.2    & 64.7    &  54.9   & 58.1               \\ 
 &  & ours (top-4)                              &                         & \textbf{157.6}   & \textbf{61.1}    &\textbf{ 65.1}    & \textbf{64.2}    & \textbf{69.0}    & \textbf{58.0}     &    \textbf{61.3}   \\ 
 &  & upper                                    &                         & 166.0   & 72.5    & 77.0    & 67.7    & 72.6    & 77.1  & 81.1       \\ \hline
\multirow{10}{*}{LF-Net~\cite{ono2018lf}} &       \multirow{10}{*}{LF-Net~\cite{ono2018lf}}      
    & lower                                    &  \multirow{5}{*}{128.0} & 45.2   & 27.9   & 28.5    &  37.1   & 37.8    & 19.1    & 19.6                              \\
 &  & LF-Net                                  &                          & 44.2   & 22.2    & 22.6    & 29.7    &  30.2   & 14.9     & 15.3                         \\
 &  & ours                                    &                          & 42.6   & 35.6    & 36.6    & 42.1    & 43.1    & 29.4     & 30.2          \\ 
 &  & ours (top-4)                              &                         & \textbf{44.3}   & \textbf{44.4}    & \textbf{45.6}    & \textbf{51.6}    & \textbf{52.8}    & \textbf{37.4}     & \textbf{38.6}      \\ 
 &  & upper                                    &                         & 46.1   & 60.9   & 63.0    & 60.1    & 62.1    & 61.7    & 63.8     \\  \cline{3-11}
 &  & lower                                    &  \multirow{5}{*}{512.0}  & 164.4   & 27.0     & 27.9    & 35.7    & 36.7    & 18.6     &    19.3                      \\
 &  & LF-Net                                  &                          & 160.3   & 21.0     & 21.7    & 28.2    & 29.1    & 14.0     &    14.6                      \\
 &  & ours                                    &                          & 155.0   & 37.1     & 38.8    & 43.6    & 45.5    & 30.7     &    32.3       \\ 
 &  & ours (top-4)                              &                         & \textbf{168.5}   & \textbf{46.6}     & \textbf{48.7}    & \textbf{54.0}    & \textbf{56.2}    & \textbf{39.4}     &    \textbf{41.4}   \\ 
 &  & upper                                    &                         & 184.6   & 66.4     & 69.6    & 64.4    & 67.4    & 68.3     &    71.7    \\ \hline
\multirow{10}{*}{RF-Net~\cite{shen2019rf}} &        \multirow{10}{*}{RF-Net~\cite{shen2019rf}}  
    & lower                                    &  \multirow{5}{*}{127.6} & 59.7   & 27.7    & 29.5    & 38.0    & 39.9    & 17.9     & 19.5                         \\
 &  & RF-Net                                    &                        & \textbf{70.4}   & 35.5    & 45.1    & 52.6    & \textbf{64.3}    & 19.1     & 26.5                         \\
 &  & ours                                    &                          & 66.5   & 41.6    & 50.6    & 49.9    & 57.3    & 33.6     & 44.1          \\ 
 &  & ours (top-4)                              &                         & 66.7   & \textbf{45.0}    & \textbf{55.7}    & \textbf{53.5}    & 62.5    & \textbf{36.7}     & \textbf{49.2}      \\ 
 &  & upper                                    &                         & 65.2   & 48.9    & 61.1    & 55.2    & 64.9    & 42.9     & 57.5    \\  \cline{3-11}
 &  & lower                                    &  \multirow{5}{*}{510.4} & 217.2   & 27.9     & 29.3    & 38.8    & 40.3    & 17.4     &    18.6        \\
 &  & RF-Net                                    &                        & \textbf{235.5}   & 37.6     & 47.8    & 56.8    & \textbf{68.3}    & 19.1     &    28.1                      \\
 &  & ours                                    &                          & 226.7   & 45.7     & 54.9    & 54.2    & 61.2    & 37.5     &    48.8       \\ 
 &  & ours (top-4)                              &                         & 230.9   & \textbf{49.6}     & \textbf{60.0}    & \textbf{58.3}    & 66.1    & \textbf{41.2}     &    \textbf{54.1}   \\ 
 &  & upper                                    &                         & 225.1   & 55.3     & 67.0    & 60.5    & 69.0    & 50.4     &    65.1    \\ \hline
\multirow{10}{*}{Key.Net~\cite{barroso2019key}} & \multirow{10}{*}{HardNet~\cite{mishchuk2017working}}  
    & lower                                    &  \multirow{5}{*}{125.5}  & 46.5   & 55.7    & 57.1    & 62.8    & 64.0    & 48.8     & 50.3                         \\
 &  & Key.Net                                 &                          & \textbf{58.9}   & 70.6    & 74.3    & \textbf{73.6}    & \textbf{77.3}    & 67.6     & 71.4                         \\
 &  & ours                                    &                          & 55.6   & 66.2    & 69.5    & 68.9    & 72.0    & 63.7     & 67.0          \\
 &  & ours (top-9)                              &                         & \textbf{58.9}   & \textbf{71.1}    & \textbf{74.9}    & 73.2    & 76.8    & \textbf{69.0}     & \textbf{73.1}      \\ 
 &  & upper                                    &                         & 47.6   & 75.6    & 78.3    & 72.3    & 74.9    & 78.7     & 81.5    \\  \cline{3-11}
 &  & lower                                    & \multirow{5}{*}{503.9}  & 177.9   & 54.7     & 56.1    & 60.7    & 62.0    & 48.9     &    50.3                      \\
 &  & Key.Net                                 &                          & \textbf{241.5}   & 73.0     & 77.9    & \textbf{74.5}    & \textbf{80.0}    & 71.5     &    75.9                      \\
 &  & ours                                    &                          & 221.5   & 68.6     & 72.8    & 69.4    & 73.8    & 67.9     &    71.8       \\ 
 &  & ours (top-9)                              &                         & 240.7   & \textbf{73.3}     & \textbf{78.2}    & 74.0    & 79.2    & \textbf{72.7}     &    \textbf{77.2}   \\ 
 &  & upper                                    &                         & 187.2   & 78.4     & 81.7    & 74.2    & 77.4    & 82.5     &    85.8    \\ \hline
\end{tabular}
}
\caption{
Additional results with off-the-shelf keypoints detectors and descriptors on the HPatches. 
Column `K' denotes the number of extracted keypoints, and `M' denotes the average number of predicted matches. 
Row with `ours' means \texttt{argmax} selection, and row with `ours (top-k)' is $k$ multiple candidates extraction on our histogram representation. 
Row with `lower' means lower bound accuracy with no patch alignment, identity patch sampling.
Row with `upper' means upper bound accuracy with patch extraction using ground-truth homography matrix.
}
\label{tab:matching_both}
\end{table}


\section{Analysis on prediction ranges.}\label{sec:supp_pred_range}
\begin{figure}[t!]
    \begin{center}
    \scalebox{0.4}{
    \includegraphics{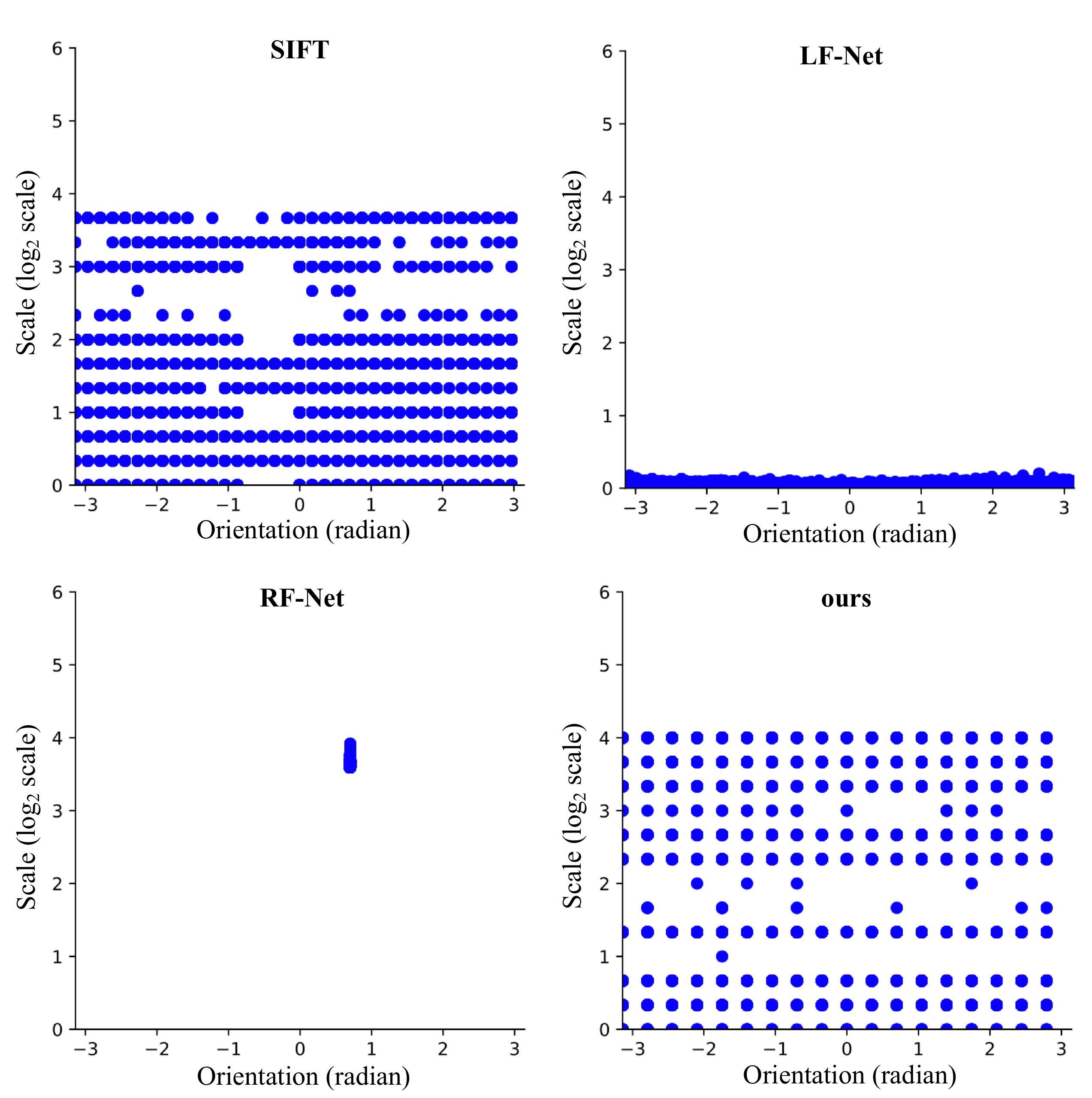}
    }
    \end{center}
    \caption{Plot of scale and orientation predictions on the PatchPose dataset. 
    The objective of this experiment is to show the limitation of the range of scale and orientation predicted by each model.
    SIFT~\cite{lowe2004distinctive} and our model have a large range in scale and orientation prediction, but LF-Net~\cite{ono2018lf} has a restricted range in scale prediction, and RF-Net~\cite{shen2019rf} has a small range in scale and orientation prediction. 
    }
    \label{fig:prediction_range}
\end{figure}

\noindent \textbf{A plot of prediction range. }
Figure~\ref{fig:prediction_range} plots the predicted scale and orientation of different models on the PatchPose dataset, where we can observe the ranges of predicted scale and orientation. 
We obtain the scale value of SIFT~\cite{lowe2004distinctive} using the scale-space maxima and octave index, where a single octave-level difference is the same to enlarge twice the image size. We obtain the orientation value of SIFT~\cite{lowe2004distinctive} using the dominant direction of the histogram bin.
We use the scale and orientation values from the output of LF-Net~\cite{ono2018lf} and RF-Net~\cite{shen2019rf} keypoints detector. In the case of our model, we convert histogram bins to the corresponding scale and orientation values by \texttt{argmax} selection. All the scale values are plotted in $\log_2$ scale.
SIFT~\cite{lowe2004distinctive} and our model show a large range in both scale prediction and orientation prediction. 
In contrast, LF-Net~\cite{ono2018lf} is limited to a small range in scale prediction, and RF-Net~\cite{shen2019rf} is to a small range in both scale prediction and orientation predictions. 

\begin{figure}[t!]\centering
{\label{fig:range_eval_sca}\includegraphics[width=.8\linewidth]{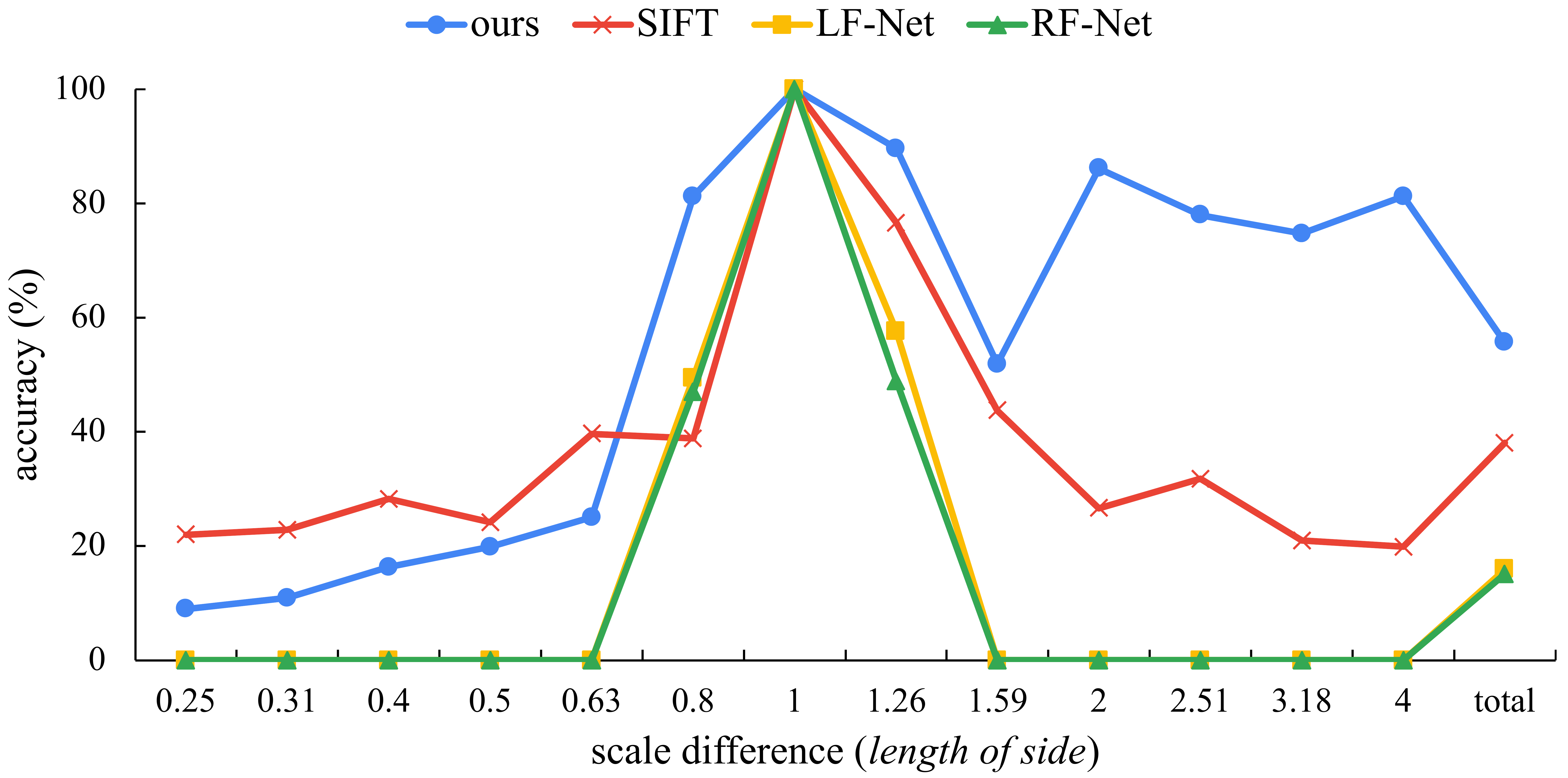}}\par \vspace{0.4cm}
{\label{fig:range_eval_ori}\includegraphics[width=.8\linewidth]{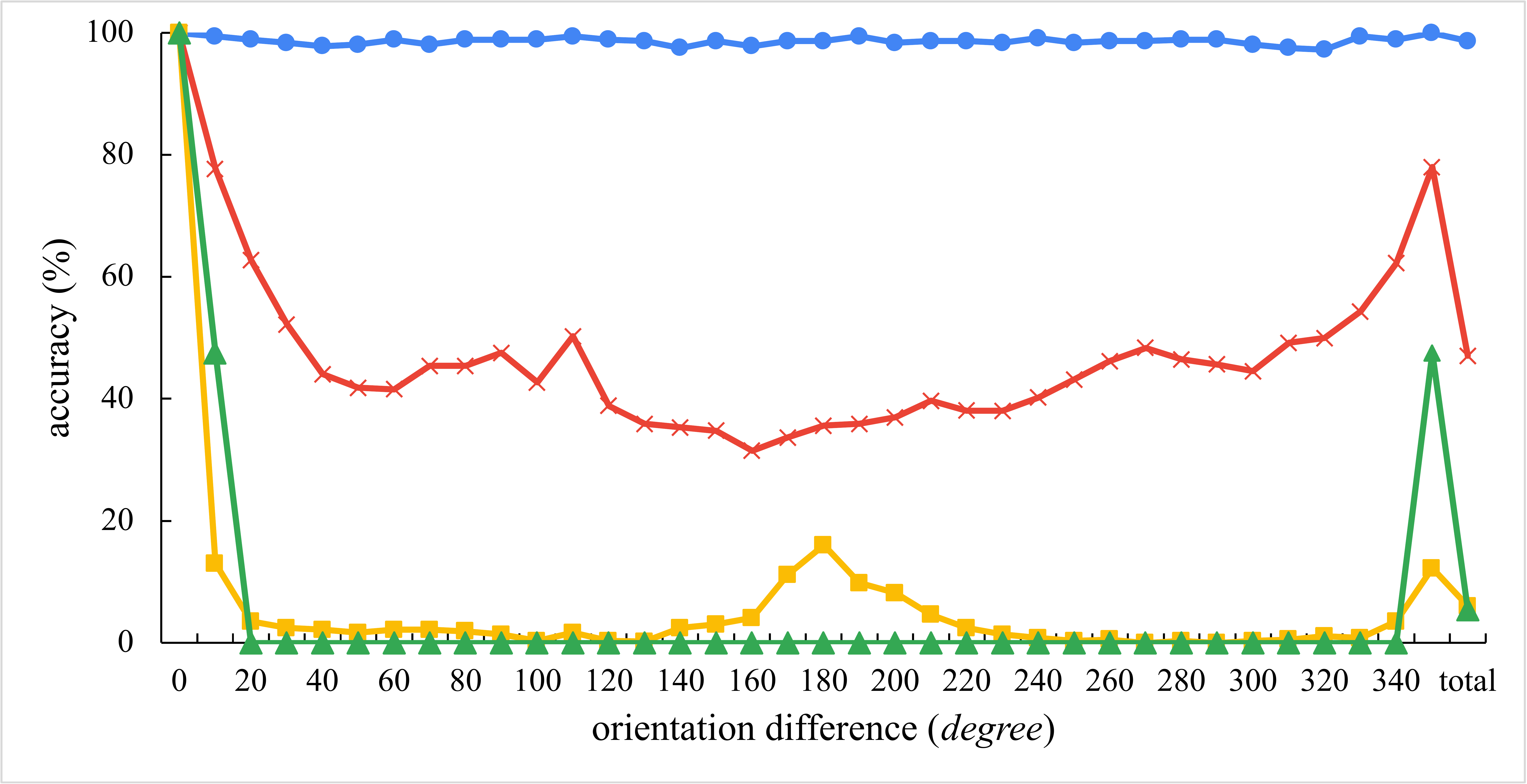}}
\caption{
Range-wise evaluation of scale estimation (upper) and orientation estimation (below) on the PatchPose dataset.
We evaluate the accuracy of scale and orientation estimation by each difference level. 
We set the accuracy threshold $\pm 2^{1/3}$ and $\pm 10 \degree$, scale and orientation, respectively.
The value of the last index `total' indicates the average of the accuracy.
}
\label{fig:rangewise_eval}
\end{figure}

\noindent \textbf{Evaluation of scale/orientation estimation under varying differences.}
Figure~\ref{fig:rangewise_eval} shows the accuracy of patch pose estimation under varying differences in scale or orientation.
Among the baselines, SIFT~\cite{lowe2004distinctive} makes relatively accurate predictions across various ranges of scale/orientation differences.
On the other hand, the existing learning-based methods~\cite{ono2018lf, shen2019rf} exhibits high accuracy only in specific ranges of scale/orientation differences.
As can be in Figure~\ref{fig:prediction_range}, this problem occurs due to the limited prediction range. 
This indicates that these models~\cite{ono2018lf, shen2019rf} have limited power in predicting large changes in scale/orientation.
In contrast, our model makes more diverse and more accurate predictions than the other methods.


\section{More qualitative results}\label{sec:supp_qual}

\begin{figure}[t]
    \centering
    \scalebox{0.72}{
    \includegraphics[]{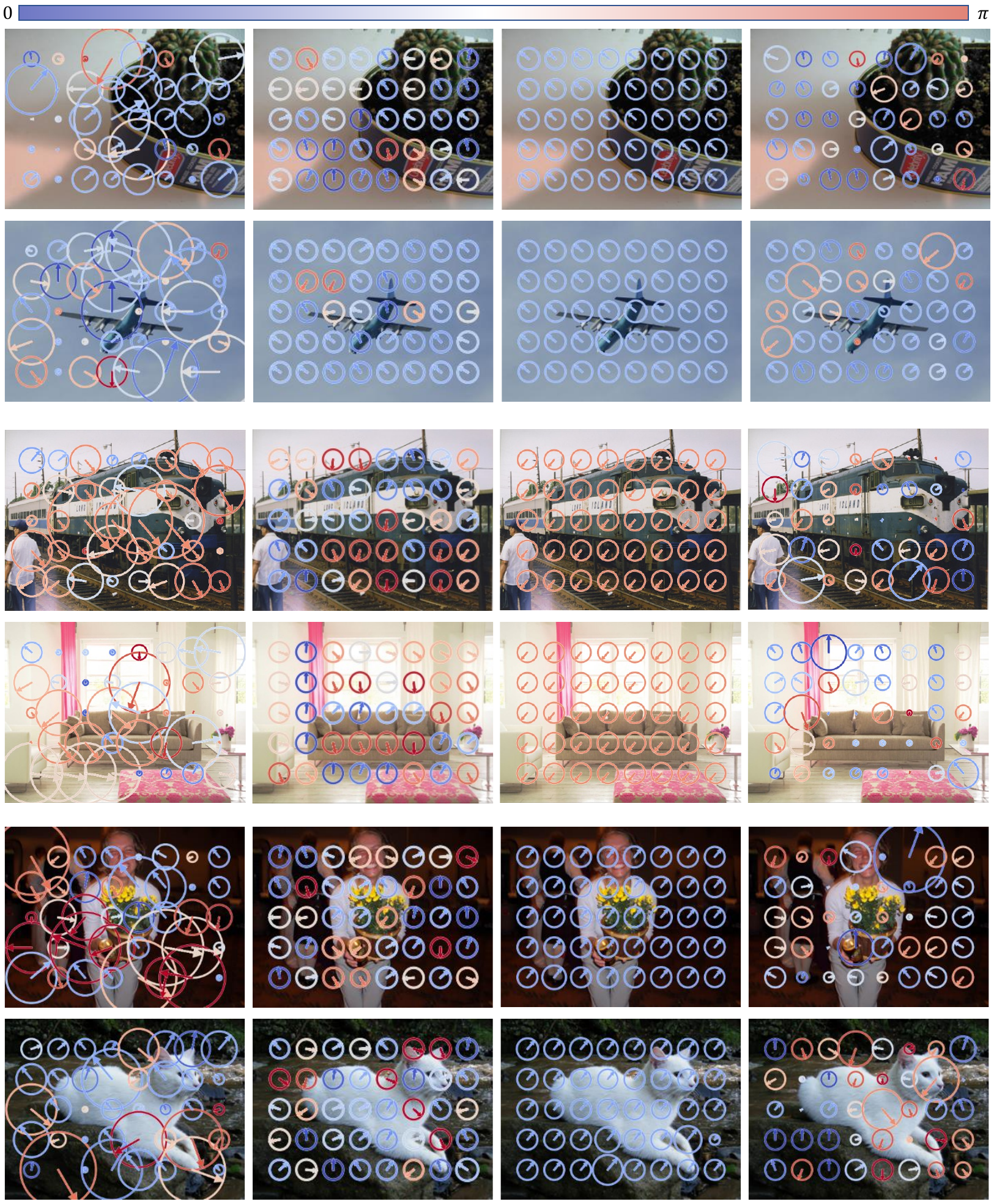}
    }
    \caption{Visualization of scale and orientation prediction error using~\cite{lowe2004distinctive, ono2018lf, shen2019rf} and ours. 
    For each image, we generate an image pair using 2.52 times upscale and  $50\degree$ rotation (Row 1, 2), $140\degree$ rotation (Row 3, 4), and $320\degree$  rotation (Row 5, 6). 
    We apply each method to the pair and estimate the difference of scale and orientation for corresponding regions. 
    The line direction denotes the orientation error, \ie, pointing upwards shows 100\% accuracy while pointing downward shows 0\% accuracy.
    For enhanced visibility, we add colors to the circles, as the error distribution at the top of the color bar. The redder the color, the less accurate the orientation predictions are; the bluer the color, the more accurate.
    The circle size represents the scale prediction error, where the larger circle size denotes the larger prediction error.
    }
    \label{fig:final_vis}
\end{figure}

We visualize the scale and orientation estimation on example cases with large pose differences in Figure~\ref{fig:final_vis}.
Each model is tested on the local image patches centered on the circles shown. 
Compared to the other methods, our method predicts the scale and orientation of the local region more accurately. 
In these challenging cases with extreme differences, the previous learning-based methods, LF-Net~\cite{ono2018lf} and RF-Net~\cite{shen2019rf}, fail to estimate large scale/orientation changes accurately, which are rarely observed in their training.
The hand-crafted method, SIFT~\cite{lowe2004distinctive}, performs relatively better than the learning-based methods. 

Figure~\ref{fig:final_vis} shows qualitative examples by different ranges. We use two more rotation angles for the experiment: $\frac{5 \pi}{18}$, $\frac{7 \pi}{9}$ and $\frac{16 \pi}{9}$. 
We first rotate each image using the target angle, then enlarge the image to be 2.52 times the original size. 
Then, we search for the corresponding locations of sampled points in the transformed image to predict the characteristic scale and orientation of corresponding locations using each method~\cite{lowe2004distinctive, ono2018lf, shen2019rf}. 
We compare the predicted pose differences with the ground truth pose differences and visualize them with colored circles and lines. 
These examples show that the previous learning-based methods tend to learn a bias in their training setup, and thus often do not generalize well to different unseen cases. 
In contrast, due to self-supervised and explicit learning, our method performs very robustly to such a wide range of variations. 

\begin{figure}[t]
\scalebox{0.5}{
    \centering
    \includegraphics{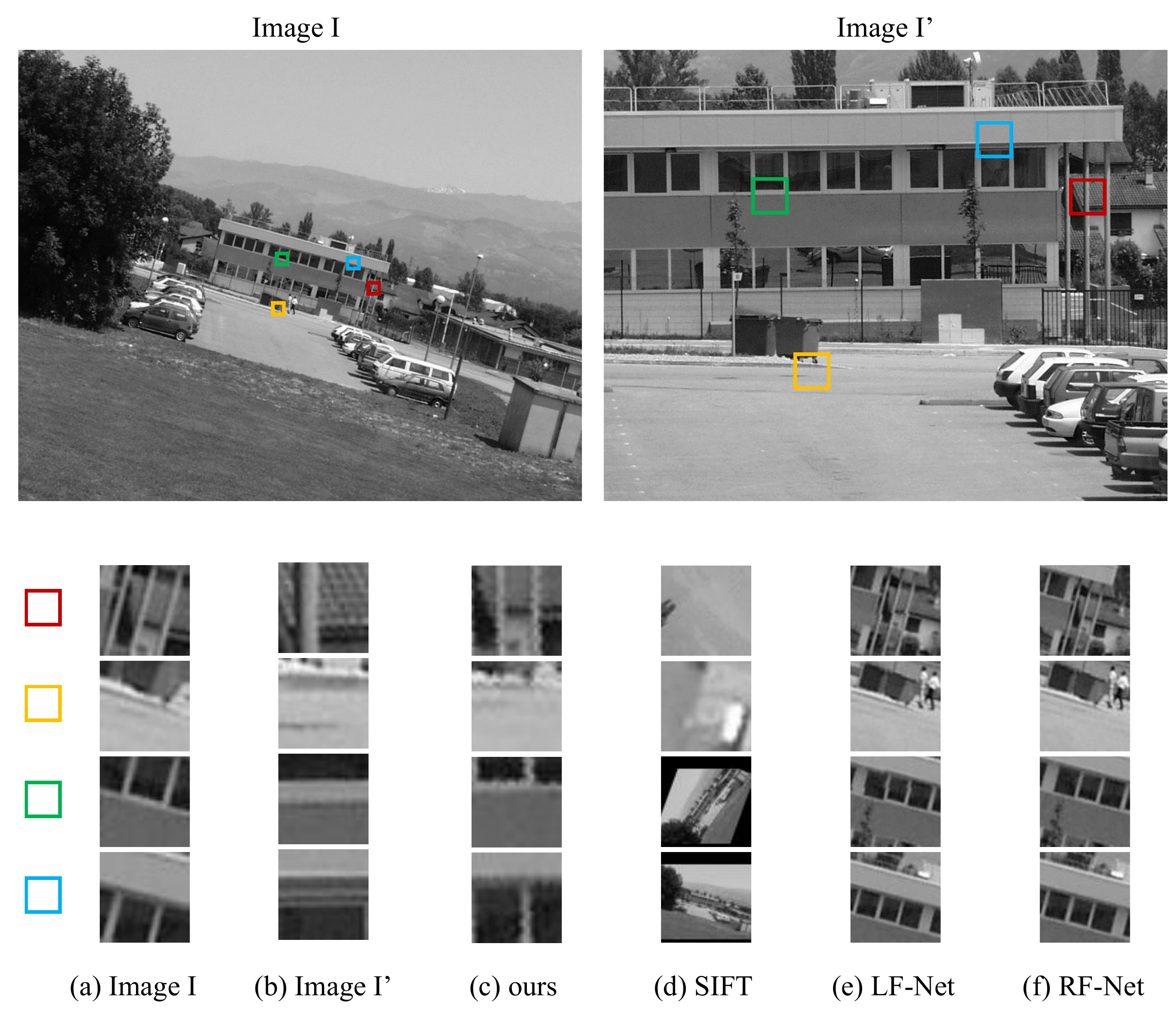}
    }
    \caption{Selected visualization of patch pose estimation on zoom/rotated image pair~\cite{mikolajczyk2005comparison, mikolajczyk2005performance}. The colors denote the corresponding patches on image $I$ and $I'$. Columns (a) and (b) show the cropped patches from the input image $I$ and $I'$. Columns (c), (d), (e), and (f) show the aligned patches from $I$ to $I'$ using the predicted scale/orientation values of each model. In this dataset, the ground-truth homography has not only scale and orientation factors but also other perspective transformation parameters ({\em e.g.,} shearing, tilting). Therefore, the patches may not be perfectly aligned from $I$ to $I'$ by only using scale and orientation factors. Nevertheless, our model consistently aligns several patches better than the other models.  } 
    \label{fig:vis_zr_seq}
\end{figure}

Figure~\ref{fig:vis_zr_seq} shows qualitative examples on zoom/rotated image pair~\cite{mikolajczyk2005comparison, mikolajczyk2005performance}. Compared to the existing models~\cite{lowe2004distinctive, ono2018lf, shen2019rf}, our model shows better patch alignment results based on the predicted scale/orientation values. 

\begin{figure*}
    \centering
    \scalebox{0.55}{
    \includegraphics{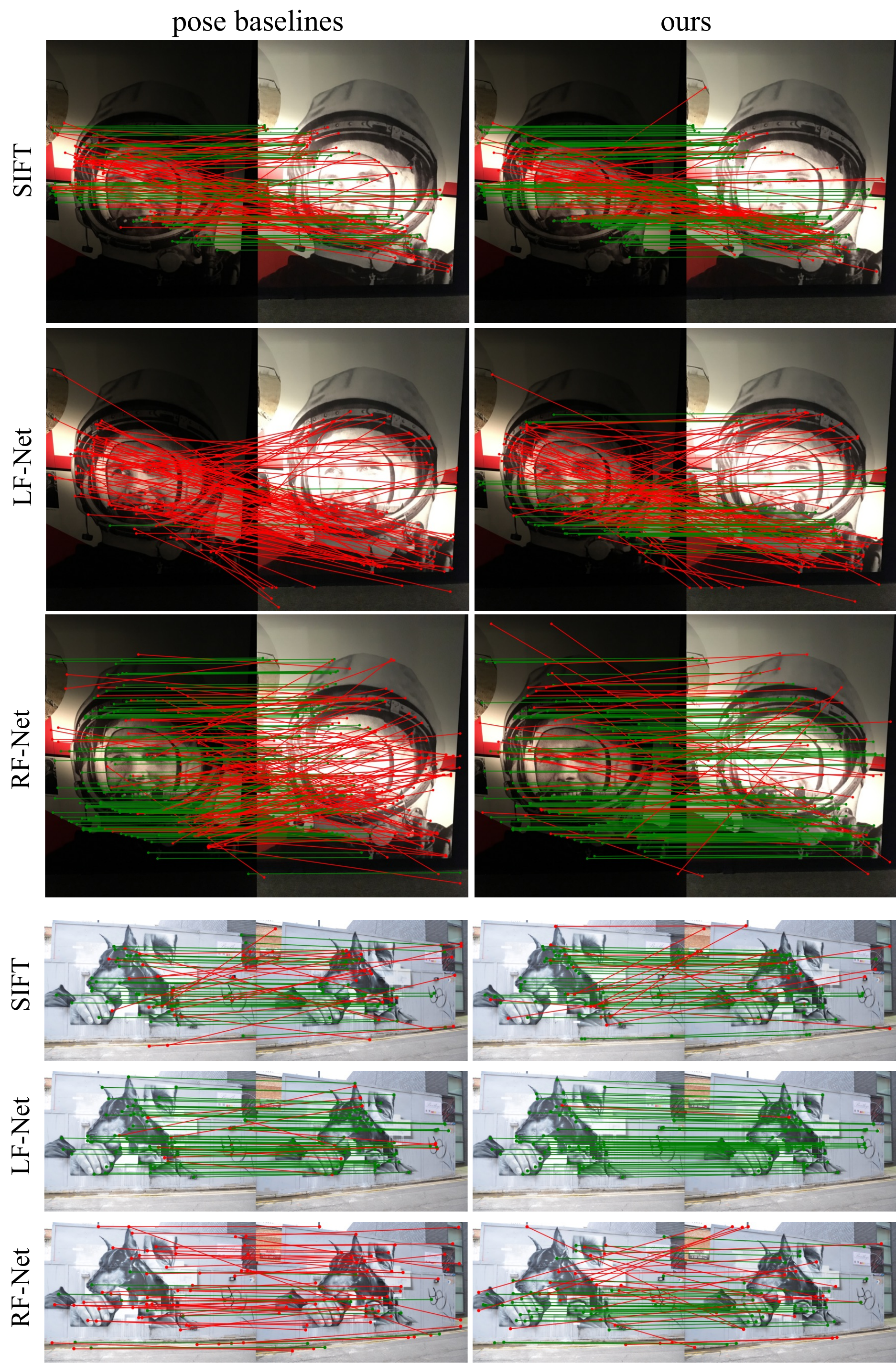}
    }
    \caption{Visualization of image matching using patch extraction on the HPatches~\cite{balntas2017hpatches}. 
    The circles are keypoints detected by each method, and the lines denote predicted matches to satisfy mutual nearest neighbour.
    The color denotes correctness by ground-truth homography, \textit{i.e.,} green color is a correct, red color is an incorrect match. 
    We set the threshold of correctness as 10 pixels. 
    The left side image pairs denote image matching results using the internal pose estimator of each method.
    The right side image pairs denote image matching results to replace the pose estimation results with our pose estimation results.
    }
    \label{fig:vis_matching_hpatches}
\end{figure*}

Figure~\ref{fig:vis_matching_hpatches} visualizes examples of image matching on the HPatches dataset~\cite{balntas2017hpatches}.
Our model generates more matches with better precision on the examples.

\begin{figure*}
    \begin{center}
    \scalebox{0.6}{
    \includegraphics{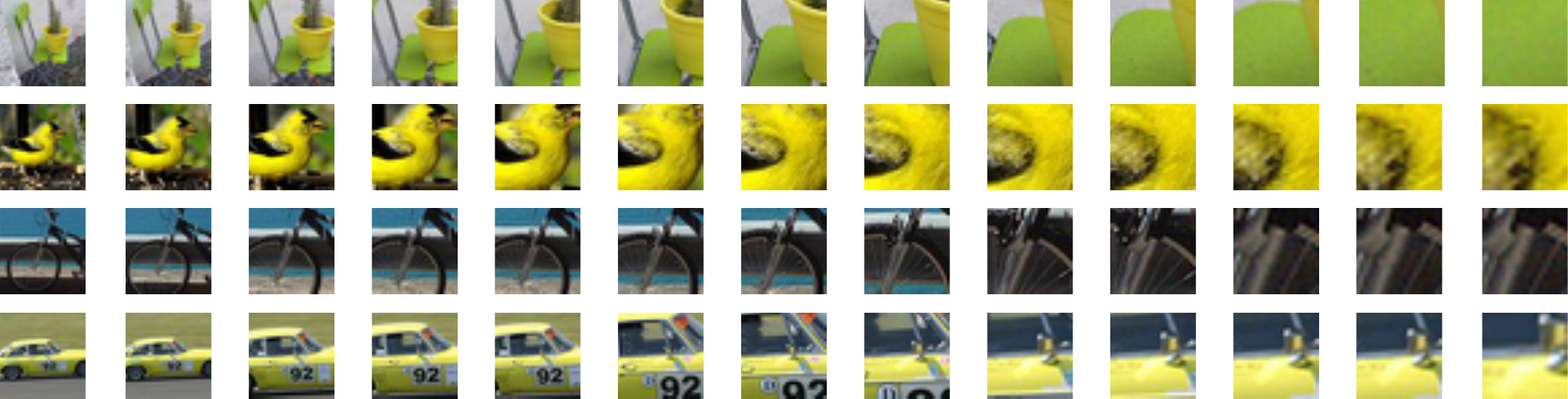}
    }
    \end{center}\vspace{-0.5cm}
    \caption{Visualization of scale variation examples of the PatchPose-A (row 1, 2) and the PatchPose-B (row 3, 4). From left to right, the columns show the resized patches which are scaled at factors from $2^{-2}$ to $2^{2}$ at an interval of $2^{\frac{1}{3}}$ about rows 1 and 2. The patches of rows 3 and 4 unlock the   interval restriction and randomly resized on the range of factors [$2^{-2}$, $2^{2}$].}
    \label{fig:vis_patchpose_scale}
\end{figure*}

\begin{figure*}
    \begin{center}
    \scalebox{0.6}{
    \includegraphics{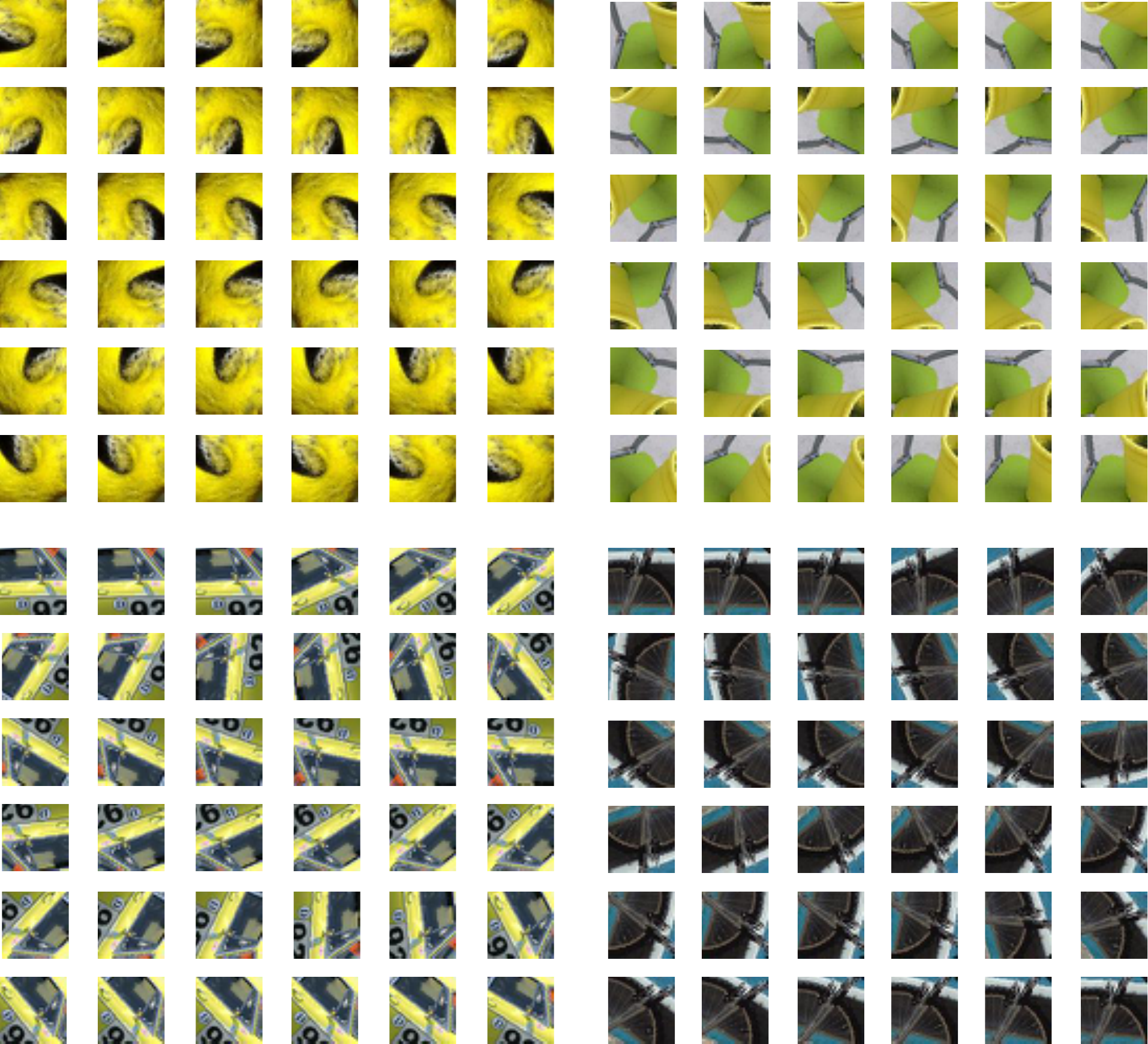}
    }
    \end{center}\vspace{-0.5cm}
    \caption{Visualization of orientation variation examples of the PatchPose-A (the top two 6x6 grids) and the PatchPose-B (the bottom two 6x6 grids). The top two grids show the rotated patches of PatchPose-A, at a rotation increasing anticlockwise from top left to bottom right at a $\frac{1}{18}\pi$ interval. The bottom two blocks show the rotated patches of PatchPose-B, at randomly generated rotation factors $0 \leq \Delta_o < 2\pi , \Delta_o \in \mathbb{R}$. We sort the sequences in an increasing order of degrees.}
    \label{fig:vis_patchpose_orient}
\end{figure*}

Figures~\ref{fig:vis_patchpose_scale} and~\ref{fig:vis_patchpose_orient} show example sequences of the PatchPose-A and the PatchPose-B. All the visualized patches are cropped by $64 \times 64$ size from the original images. 
PatchPose-A has grid-level scale/orientation variations, while the PatchPose-B has randomly selected, continuous scale/orientation variations.




\clearpage

\bibliography{egbib}